\documentclass[letterpaper,twocolumn,10pt]{article}
\usepackage{usenix}

\usepackage{tikz}
\usepackage{amsmath}
\usepackage{subfigure}
\usepackage{fmtcount}
\usepackage{graphicx}
\usepackage{xspace}
\usepackage{url}
\usepackage{acronym}
\usepackage{adjustbox}
\usepackage{latexsym}
\usepackage{courier}
\usepackage{xcolor}
\usepackage{comment}
\usepackage{longtable}
\usepackage{tabularx}
\usepackage{amssymb}
\usepackage{array}
\usepackage{multirow}

\begin{document}

\date{}

\title{Rethinking Machine Learning Robustness \\ via its Link with the Out-of-Distribution Problem}

 \author{
 {\rm Abderrahmen Amich}\\
 University of Michigan-Dearborn\\
 aamich@umich.edu
 \and
 {\rm Birhanu Eshete}\\
 University of Michigan-Dearborn\\
 birhanu@umich.edu
 } 

\maketitle

\begin{abstract}

Despite multiple efforts made towards robust machine learning (ML) models, their vulnerability to adversarial examples remains a challenging problem ---which calls for rethinking the defense strategy. In this paper, we take a step back and investigate the \textit{causes} behind ML models' susceptibility to adversarial examples. In particular, we focus on exploring the cause-effect link between adversarial examples and the out-of-distribution (OOD) problem. To that end, we propose an OOD generalization method that stands against both adversary-induced and natural distribution shifts. Through an {\em OOD to in-distribution mapping} intuition, our approach {\em translates} OOD inputs to the data distribution used to train and test the model. Through extensive experiments on three benchmark image datasets of different scales (MNIST, CIFAR10, and ImageNet) and by leveraging image-to-image translation methods, we confirm that the adversarial examples problem is a special case of the wider OOD generalization problem. Across all datasets, we  show that our translation-based approach consistently improves robustness to OOD adversarial inputs and outperforms state-of-the-art defenses by a significant margin, while preserving the \textit{exact} accuracy on benign (in-distribution) data. Furthermore, our method generalizes on naturally OOD inputs such as darker or sharper images.
\end{abstract}

\section{Introduction}\label{sec: intro}
Machine learning (ML) has shown impressive leaps in solving predictive tasks in domains such as image classification, natural language processing, voice command processing, healthcare, and malware detection. Despite their demonstrated human-surpassing capacity in experimental settings, ML models struggle to maintain their accuracy when deployed in practice, specifically, when facing inputs that lie {\em out of distribution} of their train/test data. The main reason for the poor performance of ML models in practice has to do with the fundamental assumptions that govern model training and testing: the IID assumptions~\cite{goodfellow2019research}. Under the IID assumptions, training and test examples are all generated \textit{independently} and from an \textit{identical} probability distribution. Given a training data and a test data, both of which are drawn IID from an identical distribution, the goal of the training is to learn a model that generalizes well on test data, i.e., \textit{in-distribution generalization}.

When a ML model is deployed in the wild, such as in a ML-as-a-Service (MLaaS) setting, inputs to the model may no more respect the IID assumptions. Under such a situation, to be useful, the model needs to generalize on inputs that lie far away from the model's natural distribution, i.e., {\em out of distribution (OOD) generalization}. Otherwise, the model's accuracy degrades to a level that makes it \textit{unreliable} for real-life deployment. Relating to ML unreliability in practice, prior works have revealed that ML models are vulnerable to \textit{adversarial examples}~\cite{amich2021morphence,FGSM,PGSM,CW} that successfully deplete prediction accuracy. 
 In this paper, we make an observation that the \textit{adversarial examples problem} is in fact a part of the wider \textit{ OOD generalization problem}, following the intuition that adversarial inputs may not conform to the IID setting due to adversarial perturbations. 

Though not all OOD samples are adversarial, we observe that most adversarial examples happen to be OOD samples~\cite{goodfellow2019research}. Through a preliminary study that compares distribution ($\mathbb{P}_{adv}$) of adversarial examples with distribution ($\mathbb{P}_{train}$) of training data, we find that adversarial examples cause distribution shift that results in fooling the model to make incorrect predictions. To be concrete, using an OOD detector on adversarial examples (described in \ref{SSD}), our preliminary experiments on CIFAR10 \cite{cifar} suggest that $90\%$ of Fast Gradient Sign Method (FGSM) \cite{FGSM} test samples and $87\%$ of Projected Gradient Descent (PGD) \cite{PGSM} test examples are OOD, confirming the lack of OOD generalization as a potential major cause of the test-time vulnerability of ML models to adversarial examples.
Recognizing the {\em cause-effect} link between the adversarial examples problem (effect) and the OOD generalization problem (cause), in this work, we aim to rethink our current approaches to defend against adversarial examples, by focusing on the wider OOD generalization problem.

Early attempts ~\cite{Early-Defense14,Early-Defense15} to harden ML models against adversarial examples provided only marginal robustness improvements. Heuristic defenses based on approaches such as {\em defensive distillation}~\cite{distillation}, {\em data transformation}~\cite{Compression17,Compression18,Augmentation17,Cropping17,Rand15,Rand18}, and {\em gradient masking}~\cite{Thermo-Encode18,PixelDefend18} were subsequently broken~\cite{CW, Carlini-Breaking17,Carlini-BreakingUsenix17,Gradient-Masking18}.
Although {\em adversarial training}~\cite{EnsembelAdvTrain18} remains effective against known attacks, robustness comes at a cost of accuracy penalty on clean data. \textit{Certified defenses}~\cite{lecuyer2019certified,RandomSmoothing19,Certified-AdditiveNoise19} are limited in the magnitude of robustness guarantee they offer and operate on a restricted class of attacks constrained to LP-norms \cite{lecuyer2019certified,wong2018provable}. Recently, {\em moving target defenses}~\cite{amich2021morphence} have emerged as promising steps to thwart repeated and correlated attacks that take advantage of the fixed target nature of models. A common thread in the current adversarial example defense literature is that, the focus has been more on answering the question \textit{``how to defend against adversarial examples?''} than exploring the question \textit{``why ML models are vulnerable to adversarial examples?''}. We argue that tackling the cause, i.e., lack of OOD generalization, is essential towards building ML models robust to OOD inputs, both adversarial and benign. In this vein, we acknowledge \cite{Features-not-Bugs19} which pointed to empirical evidence that models pick spurious correlations instead of robust ones that stand manipulations.

\begin{figure}[t!]
    
    \centering
    \includegraphics[width=\columnwidth]{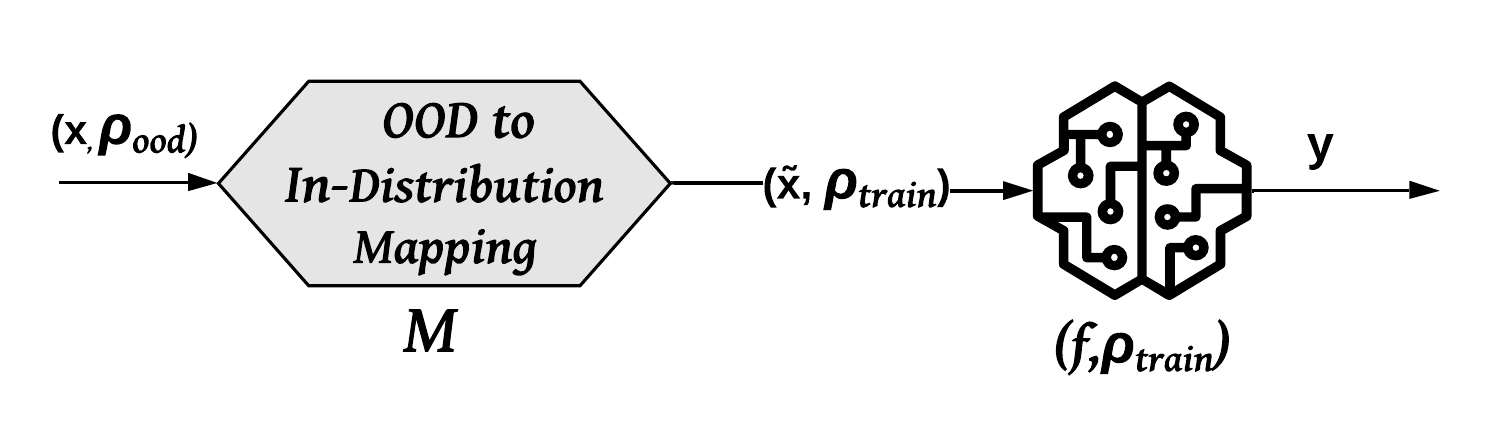}
    \vspace*{-3em}
    \caption{High-level approach intuition. $\Tilde{x} = \mathcal{M}(x)$ is the in-distribution equivalent of an OOD sample $x$ obtained after passing through the OOD to in-distribution mapping module.}
    \vspace*{-1em}
    \label{fig:framework}
\end{figure}

In this paper, we propose an approach that enables a model trained under the IID assumptions and generalizes well on in-distribution test data to also generalize well on OOD data, both adversarial and benign. Given an OOD input ($x,\mathbb{P}_{ood}$) and a model $f$ trained with the IID assumptions on a training data $X_{train}$ sampled from distribution $\mathbb{P}_{train}$, our approach intuition, illustrated in Figure \ref{fig:framework}, performs {\em OOD to in-distribution mapping} of the OOD sample $x$ drawn from $\mathbb{P}_{ood}$ to the training distribution $\mathbb{P}_{train}$ of the model. We denote the OOD to in-distribution mapping module as $\mathcal{M}$ and the act of mapping an OOD sample $x$ to its in-distribution equivalent as $\Tilde{x} = \mathcal{M}(x)$. Although the notion of OOD to in-distribution mapping is adaptable to other domains/tasks, in this work we focus on a widely practical ML task of image classification.

To implement $\mathcal{M}$, we specifically leverage recent advances in {\em Image-to-Image translation}~\cite{CycleGAN2017,isola2018imagetoimage} that aim to transfer images from a source distribution to a target distribution while preserving content representations. To this end, we explore alternative designs of $\mathcal{M}$ to map an OOD sample to its in-distribution equivalent that is correctly classified by $f$ (hence $f$ is much more robust to benign or adversary-induced distributional shifts). In Section \ref{sec: approach}, we expand Figure \ref{fig:framework} to thoroughly describe the components of our approach, including the need for an OOD detector to put apart OOD and IID inputs. We note that previous efforts in the OOD generalization literature may as well be leveraged to serve the purpose of our approach. However, in this work, we choose \textit{OOD to in-distribution mapping of the input} instead of \textit{changing the model itself} so as to avoid potential tampering with a model's ability to generalize on in-distribution data. 

We extensively evaluate our approach across multiple benchmark datasets (MNIST, CIFAR10, ImageNet), multiple reference attacks that generate OOD adversarial examples (FGSM ~\cite{FGSM}, PGD~\cite{PGSM}, C\&W~\cite{CW}, SPSA~\cite{uesato2018adversarial}), and alternative designs of $\mathcal{M}$. When deployed as defense, our OOD generalization-based approach outperforms prior defense techniques by a margin of $\approx 27.64\%$ on MNIST, $\approx 40.25\%$ on CIFAR10, and $\approx 40.23\%$ on ImageNet, averaged across all studied attacks (i.e., FGSM, PGD, C\&W, SPSA). 
All this is achieved while preserving the \textit{exact} original accuracy on benign data. Furthermore, it generalizes well on non-adversarial OOD examples. Specifically, it improves the accuracy of a CIFAR10 model on OOD \textit{darker images} ($\mathbb{P}_{ood}=\mathbb{P}_{dark}$) by a margin of $35.64\%$ and increases the accuracy of an ImageNet model on OOD \textit{sharper images} ($\mathbb{P}_{ood}=\mathbb{P}_{sharp}$) by a margin of $34.16\%$.

In summary, we make the following main contributions:
\begin{enumerate}
\item  We empirically illustrate that the adversarial examples problem is a special case of the OOD generalization problem.
\item We propose an OOD Generalization approach able to defend against adversarial examples much better than \textit{ensemble adversarial training} \cite{EnsembelAdvTrain18} and \textit{adversarial training} \cite{salman2020adversarially} on MNIST, CIFAR10, and ImageNet, while also generalizing on non-adversarial OOD inputs.
\item  We propose the first defense that preserves \textit{the exact} accuracy on benign data.

\item We leverage the utility of Image-to-Image translation methods to overcome the OOD generalization problem in different case studies (e.g., adversarial examples, benign OOD examples).
\item Our code is available at \url{https://github.com/OOD2IID/code}.
\end{enumerate}

\section{Background}\label{sec: bground}


\subsection{Generative Adversarial Networks}

First introduced by Goodfellow et al.~\cite{goodfellow2014generative}, generative adversarial networks (GANs) are a deep-learning-based generative models, that have proved their capability to generate fake data, similar to real training data. It is a framework in which two neural networks, a generator $G$ and a discriminator $D$, contest each other in a minimax two-player game. The generator aims to trick the discriminator by presenting it fake examples (e.g., generated samples), that can be classified as real. The goal of the discriminator is to identify whether the samples coming from $G$ are generated (fake) or real. With respect to an adversarial loss the generator is trained to fool the discriminator. Then improved architectures have been proposed to produce more stable models (DCGAN\cite{radford2016unsupervised}) and conditional GANs where the generator is trained subject to a condition given by an additional input other than the random vector from the latent space \cite{mirza2014conditional}. The additional input could be a class value (e.g., a digit for MNIST). It can be used to generate a fake image of a specific label.

\vspace*{-1em}
\subsection{Image-to-Image Translation}\label{subsec:I2I}

\textbf{Overview:} \textit{Image-to-Image Translation} is a vision and graphics problem where the goal is to map an input image to an output image that represents its translation to a different distribution \cite{CycleGAN2017}. After the emergence of GANs, several approaches used GANs to build accurate models that learn the Image-to-Image translation.
First, researchers proposed supervised approaches that require a \textit{paired} training set where each image from the source distribution \textit{($X$)} has a correspondent image in the target distribution \textit{($\Tilde{X}$)} \cite{long2015fully,isola2018imagetoimage}. One of the most notable works is \textit{pix2pix} which uses a conditional GAN (cGAN). The generation of target images (translated) is conditional on a given input image \cite{isola2018imagetoimage}. \textit{pix2pix} stands out from other approaches since it is generic and can be applied to different tasks (e.g., generating photographs from sketches \cite{sangkloy2016scribbler}, from attribute and semantic layouts \cite{karacan2016learning}). However, obtaining paired training data can be difficult and expensive for some applications where datasets are not largely available (e.g. semantic segmentation \cite{cordts2016cityscapes}). Consequently, \textit{unpaired} Image-to-Image translation ideas emerged (e.g., cycleGAN \cite{CycleGAN2017}, CoGAN \cite{liu2016coupled}, Auto-Encoding Variational Bayes \cite{liu2018unsupervised}), where the goal is to relate two data domains: $X$ and $\Tilde{X}$. CycleGAN has been proved to be the best approach \cite{CycleGAN2017}, and it builds on \textit{pix2pix} by proposing {\em cycle consistency loss} as a way of using transitivity to supervise training.

\textbf{Pix2Pix \cite{isola2018imagetoimage}:}
Pix2Pix is a fully supervised paired image-to-image translation approach and it employs a conditional GAN (cGAN) architecture that requires specifying a generator model $G : X \rightarrow \Tilde{X} $, discriminator model $D$, and model optimization procedure. Unlike, the traditional GAN, the generator $G$ does not take a point from the latent space as input. Instead, noise is provided only in form of dropout applied on several layers of the generator at both training and testing time. The generator model is a U-net that takes as an input an image from the source domain $X$ and is trained to return a translated image in the target domain $\Tilde{X}$. A U-net is similar to an encoder-decoder model, it involves down-sampling to a bottleneck and up-sampling again to an output image. The discriminator model takes an image from the source domain and an image from the target domain and distinguishes whether the image from the target domain is a real translation or a generated version (fake) of the source image.

\textbf{CycleGAN \cite{CycleGAN2017}:}
 CycleGAN is an unpaired image-to-image translation approach. It trains two generators,  $G : X \rightarrow \Tilde{X} $ and $F: \Tilde{X} \rightarrow X$, using a forward cycle-consistency loss: $F(G(x)) \approx x $ and a backward cycle-consistency loss: $ G(F(\Tilde{x})) \approx \Tilde{x} $. The cycle-consistency losses are used to further reduce the space of possible mapping functions, by eliminating the ones that are not cycle-consistent. As a result, for each image $x$ from domain $X$, the image translation cycle should be able to bring $x$ back to the original image (i.e., forward cycle consistency). 
 Similarly, $G$ and $F$ should also satisfy backward cycle consistency. 
 Additionally, two adversarial discriminators $D_X$ and $D_{\Tilde{X}}$ 
 are introduced where $D_X$ distinguishes between images $\{x\}$ and translated images $\{F(\Tilde{x})\}$ and $D_{\Tilde{X}}$ distinguishes between images $\{\Tilde{x}\}$ and translated images $\{G(x)\}$. The objective is the sum of two types of terms: \textit{adversarial losses $(L_{(GAN)})$} for matching the distribution of generated images to the data distribution in the target domain; and \textit{cycle consistency losses ($L_{(cyc)}$)} to prevent the learned mappings $G$ and $F$ from contradicting each other. Formally, the full objective is defined as follows:
\vspace*{-.75em}
\begin{equation}
\begin{split}
L(G,F,D_X,D_{\Tilde{X}}) = L_{GAN}(G,D_{\Tilde{X}},X,\Tilde{X})+ \\ L_{GAN}(F,D_X,\Tilde{X},X)+ \lambda L_{cyc}(G,F) 
\end{split}
\label{cycle-loss}
\end{equation}
where,
\vspace*{-.75em}
\begin{itemize}
\item $L_{GAN}(G,D_{\Tilde{x}},X,\Tilde{X}) = \mathbb{E}_{\Tilde{x} \sim \mathbb{P}_{data}(\Tilde{x})}[log D_{\Tilde{X}}(\Tilde{x})] \\+ \mathbb{E}_{x \sim \mathbb{P}_{data}(x)}[log (1-D_{\Tilde{X}}(G(x))]$
\vspace*{-.75em}
\item $L_{cyc}(G,F) = \mathbb{E}_{x \sim \mathbb{P}_{data}(x)}[||F(G(x)) - x||_1] \\+ \mathbb{E}_{\Tilde{x} \sim \mathbb{P}_{data}(\Tilde{x})}[||G(F(\Tilde{x})) - \Tilde{x}||_1]$
\vspace*{-.75em}
\item$\lambda$ controls the relative importance of the two objectives

\end{itemize}
\vspace*{-1em}

\subsection{SSD: Out-of-Distribution Detector}
\label{SSD}
SSD~\cite{SSD} is proposed to detect OOD data that lies far away from the training distribution of a ML model $f$. Its appealing side is that it is a self-supervised method that can reach good performance using only unlabeled data instead of fine-grained labeled data that can be hard to produce. Given unlabeled training data and using \textit{Contrastive self-supervised representation learning}, it trains a feature extractor by discriminating between individual samples (formal definition in paper \cite{SSD}). Next, the OOD detection is performed through a cluster-conditioned detection. Using k-means clustering, in-distribution data features are partitioned into $m$ clusters. Features of each cluster ($Z_{m}$) are modeled independently to calculate an outlier score of an input $x$, using the Mahalanobis distance (formal definition in paper \cite{SSD}).

\section{Related Work}\label{sec: related}

\subsection{Out-of-Distribution Generalization}\label{subsec:related-ood}
The OOD generalization literature is still emerging and spans multiple domains. A recent survey ~\cite{shen2021outofdistribution} provides a more in-depth discussion on the various OOD generalization techniques. Broadly, we find three lines of OOD generalization methods: unsupervised representation learning, supervised learning, and optimization. In the supervised learning line of work, to which our work is mostly related, ~\cite{IRM19} and its follow-up OOD generalization approaches (e.g., \cite{KruegerCJ0BZPC21,AhujaSVD20}) aim to find {\em invariant representations} from different training environments via invariant risk regularization. Li et al.~\cite{MetaLearning21} introduce a meta-learning procedure, which simulates train and test domain shift during training. Carlucci et al.~\cite{CarlucciDBCT19} learn semantic labels of images in a supervised way to jointly solve jigsaw puzzles on the same samples. 

\textbf{Our work:} In this work, we choose to focus on \textit{input in-distribution mapping} instead of prior OOD generalization-aware learning methods~\cite{Representation-Learning-Survey,zhou2021domain,Invariance18,rahimian2019distributionally}, to avoid any possible tempering with robust in-distribution representations in an attempt to learn stable representations for OOD generalization.

\vspace*{-1em}
\subsection{Defenses against Adversarial Examples}
The adversarial example arms race has resulted in a plethora of defense techniques, most of which were either broken~\cite{Carlini-Breaking17,Carlini-BreakingUsenix17,Gradient-Masking18} or offer empirical robustness. 

\textbf{Model hardening:} Early defense strategies aimed at model hardening ~\cite{Early-Defense14,Early-Defense15} were only marginally resilient to attacks. What followed next is dominated by best effort empirical defenses. {\em Defensive distillation}~\cite{distillation} that aims at defensively refining a neural network was broken by the Carlini-Wagner attack~\cite{CW}. {\em Adversarial training}~\cite{FGSM}, among the few viable defenses to date, sacrifices benign accuracy. {\em Gradient masking}~\cite{Thermo-Encode18,PixelDefend18} which obscures gradient information from a white-box adversary was later broken by Athalye et al.~\cite{Gradient-Masking18}. \textit{Certifiably robust defenses}~\cite{wong2018provable,raghunathan2020certified,lecuyer2019certified,RandomSmoothing19,Certified-AdditiveNoise19} aim for theoretically justified guarantee of robustness under attack, but they remain limited to low robustness guarantee in specific distance metric types (mostly $l_p$ norms). Recently, \textit{moving target defenses}~\cite{fMTD19,MTDeep19,EI-MTD,amich2021morphence} emerged intending to make models moving targets against adversarial queries to thwart repeated/correlated attacks, but maintaining the moving target aspect is often computationally expensive.

\textbf{Adversarial detection and input transformation:} Prior works have also suggested adversarial detection methods through statistical measurements \cite{Cohen_2020_CVPR,feinman2017detecting,Yang_Chen_Hsieh_Wang_Jordan_2020} or a secondary classifier \cite{Ma2019NICDA}. Statistical measurements are assessed to be unlikely to consistently detect adversarial examples, due to their intrinsically unperceptive nature. Using a secondary classifier to detect adversarial examples can reproduce the same vulnerability to a second round of adversarial examples that aim to fool it. Another similar group of works attempted to perform transformations on the test input to filter adversarial perturbations, through a preprocessing step using a Deep Denoising Autoencoder (DDA) \cite{sahay2018combatting,8890816}, or discrete cosine transform to filter off recessive features \cite{liu2021featurefilter}.

\textbf{Our work:} Despite the rich state of the art of various methods to mitigate adversarial example attacks, this research problem is still far from being solved. In this work, we go beyond \textit{model hardening} and \textit{input transformation}, by focusing primarily on the major causes of adversarial examples. Consequently, we look to the problem as a special case of the OOD generalization problem and we confirm that distribution shifts can be caused by adversarial perturbations. Therefore, ML models trained under the IID assumption are vulnerable by design to such inputs. Our approach is an OOD-Generalization-based defense that stands against OOD examples whether they are adversarial or natural OOD inputs.

\section{Approach}\label{sec: approach}

As motivated in Section \ref{sec: intro}, our goal is to shift the defense landscape through a solution that addresses one of the major causes of adversarial examples: {\em the OOD generalization problem}. First, we present an overview of our approach. Next, in subsequent sections, we describe details of the approach.
\vspace*{-1em}
\subsection{Overview}
We consider a model $f$ trained under the IID assumption on a training data $X_{train}$ sampled from a training distribution $\mathbb{P}_{train}$ and generalizes well on samples drawn independently from the same distribution $\mathbb{P}_{test}=\mathbb{P}_{train}$. However, given an OOD test sample ($x, \mathbb{P}_{ood}$), where $\mathbb{P}_{ood} \neq \mathbb{P}_{train}$ (or $\mathbb{P}_{ood} \neq \mathbb{P}_{test}$), $f$ is likely to produce an incorrect prediction on $x$, which makes it unsuitable for real-word and critical tasks. We denote the data distribution used to independently draw training and test samples by $\mathbb{P}_{data}=\mathbb{P}_{train}=\mathbb{P}_{test}$. Thus, an unknown test input $\mathbb{P}_{unknown}$ can be either OOD ($\mathbb{P}_{unknown}=\mathbb{P}_{OOD} \neq\mathbb{P}_{data}$) or IID ($\mathbb{P}_{unknown}=\mathbb{P}_{data}$).

\begin{figure}[t!]
    
    \centering
    \includegraphics[width=\columnwidth]{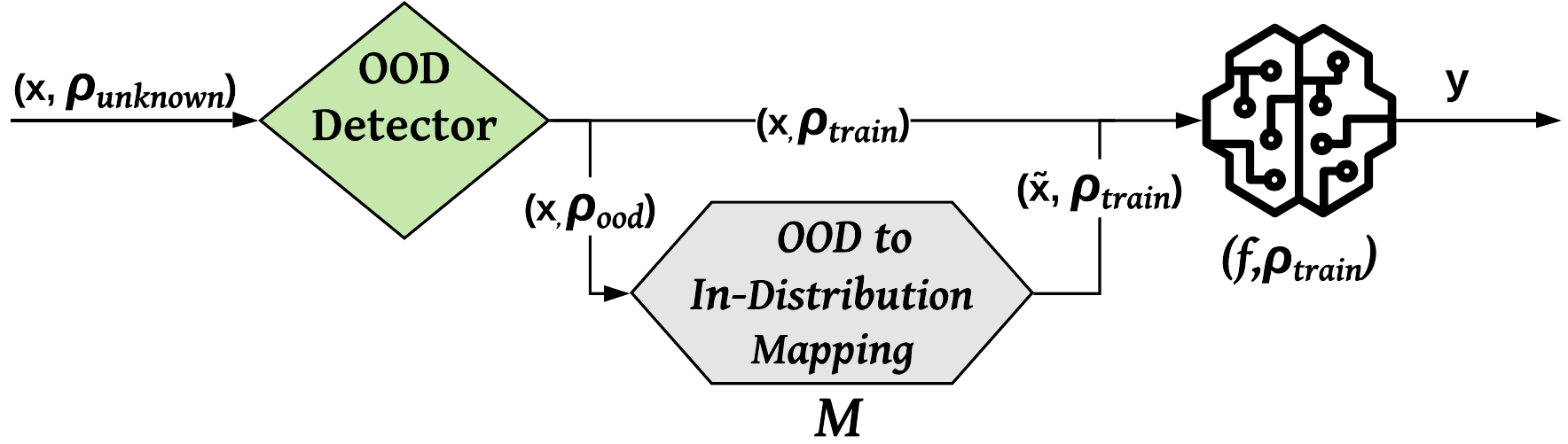}
    \vspace*{-2em}
    \caption{An overview of OOD to in-distribution mapping.}
    \vspace*{-1em}
    \label{fig:framework2}

\end{figure}

\textit{The goal of our approach is to improve the model's accuracy on OOD samples while maintaining its high accuracy generalization on in-distribution samples}.
As illustrated in Figure \ref{fig:framework2}, our approach involves two key steps: 1) detecting whether or not an input $(x,\mathbb{P}_{unknown})$ is OOD and 2) mapping an OOD input $(x,\mathbb{P}_{ood})$ to its in-distribution equivalent. In particular, a test input $(x,\mathbb{P}_{unknown}$) is first received by an \textit{OOD detector} that decides whether or not it is an OOD sample. If $(x,\mathbb{P}_{unknown})$ turns out to be an OOD sample $(x,\mathbb{P}_{ood})$, then our OOD to in-distribution mapping module $\mathcal{M}$ intervenes to convert $(x,\mathbb{P}_{unknown})$ into its in-distribution equivalent $(\Tilde{x},\mathbb{P}_{data})$ that can be correctly recognized by $f$ with a confidence as high as the model's generalization power on samples from $\mathbb{P}_{test}$. Otherwise, if $(x,\mathbb{P}_{unknown})$ turns out to be already drawn from the same data distribution $\mathbb{P}_{data}$, $f$ computes its prediction as in the IID case. 

\textbf{OOD Detection}: To rule out in-distribution samples from the OOD to in-distribution mapping goal, we leverage recent advances in OOD detection~\cite{OOD-methods}. In particular, we adopt the current state-of-the-art method called \textit{SSD} that trains a {\em self-supervised outlier detector} through learning a feature representation of $\mathbb{P}_{data}$ ~\cite{SSD} (Section \ref{SSD}). Given an input sample drawn from $\mathbb{P}_{unknown}$, SSD returns what we call an \textit{OOD score}, that reflects its distance from the in-distribution data drawn from $\mathbb{P}_{data}$. In order to effectively deploy SSD into our framework, a threshold definition ($\tau$) is necessary to separate between queries that exhibit tolerable distribution shift (i.e., $OOD score<\tau$) from those that are far away from $\mathbb{P}_{data}$ (i.e., $OOD score \ge \tau$). For instance, when our approach is deployed to defend against OOD adversarial examples (i.e., those drawn from $\mathbb{P}_{adv}$), the threshold is selected to ensure that any benign input is flagged as an in-distribution input. Consequently, it allows benign input ($x,\mathbb{P}_{test}$) to be directly classified by $f$, without considering any defense measures that could sacrifice the model's accuracy on benign data. As a result, in adversarial settings, the OOD detector plays the role of an adversarial examples detector that guides our approach to maintain the exact accuracy on benign data. In Section \ref{result:OOD_adv}, we experimentally confirm that, when deployed as a defense, our approach does not sacrifice accuracy on benign data.

\textbf{OOD to In-Distribution Mapping}: Given an OOD sample $(x,\mathbb{P}_{ood})$, the design of the OOD to in-distribution mapping module $\mathcal{M}$ is dictated by the data domain. Given the rich literature of ML and adversarial ML in the image domain and considering the high impact of the image-related systems in our modern life (e.g., healthcare, autonomous vehicles, video surveillance), we explore alternative architectures of $\mathcal{M}$ in image classification.
As stated earlier, the purpose of $\mathcal{M}$ is to translate an OOD sample ($x, \mathbb{P}_{ood}$) to its in-distribution equivalent ($\Tilde{x}, \mathbb{P}_{data}$), where $\Tilde{x} = \mathcal{M}(x)$. To this end, \textit{Image-to-Image translation} methods (Section \ref{subsec:I2I}) are naturally suitable for realizing $\mathcal{M}$, because our goal here is inline with that of translating an image from a source distribution to its equivalent in a target distribution. In the following sections, we explore design alternatives of the in-distribution translator $\mathcal{M}$. In the remainder of the paper, a translator $\mathcal{M}$ will have the form  \textbf{[Src-Dist]2[Target-Dist]}, where \textbf{Src-Dist} corresponds to source/input distribution and \textbf{Target-Dist} to target/output distribution. In the context of OOD adversarial examples, Src-Dist represents the distribution of adversarial examples crafted using attacks such as FGSM, PGD, C\&W, or SPSA. On the other hand, Target-Dist is $\mathbb{P}_{data}$ of the model at hand.
 
 In the following, we examine two alternative approaches to realize $\mathcal{M}$ with respect to realistic assumptions of a ML system deployed as a service.
 
     
\vspace*{-1em}
\subsection{Standalone Translator}\label{subsec:standalone-trans}
We first discuss the case where we consider only one translator: $\mathcal{M}=\mathcal{M}_1$ = Src-Domain2Target-Domain.
\vspace*{-1em}
\subsubsection{Source Domain = One Distribution}
A standalone translator is an in-distribution mapping model trained to solely translate inputs from a specific source distribution into $\mathbb{P}_{data}$. Considering the case of adversarial examples, where our approach is deployed as a defense, the source distribution is the distribution of adversarial examples (e.g., $\mathbb{P}_{FGSM}$, $\mathbb{P}_{PGD}$, etc). For OOD benign samples, where our approach plays the role of an OOD generalization method, the source distribution can be any other OOD distribution (i.e., $\mathbb{P}_{OOD}\neq\mathbb{P}_{data}$). For instance, blurry images ($\mathbb{P}_{blur}$) or sharper images ($\mathbb{P}_{sharp}$).

In practice, an input query may be drawn from any unknown distribution $\mathbb{P}_{unknown}$. Particularly, the deployer of our framework has no knowledge of the distribution of an input queried by a user. As a result, there is no direct approach to select the source domain to use to train the in-distribution  translator.  For instance, if deployed as a defense against adversarial examples, there is no guarantee that a standalone translator, trained to map FGSM examples into in-distribution samples (i.e., FGSM2Benign), is able to additionally translate other attack examples (e.g., PGD, C\&W). This issue is similarly faced by adversarial training techniques (e.g., \cite{EnsembelAdvTrain18}). For instance, a model trained on FGSM samples is not necessarily robust against samples from other attacks (e.g., C\&W). Hence, in addition to exhaustively examining different attacks as a source domain (Section \ref{result:OOD_adv}), in the following sections, we discuss other designs to mitigate this issue.
\vspace*{-1em}
\subsubsection{Source Domain =  Mixture of Distributions}

Intuitively, one way to enhance the translation capabilities of a standalone translator $\mathcal{M}: X \rightarrow \Tilde{X}$ on different distributions is to train it on a mixture of distributions as a source domain, i.e., source domain $X={\displaystyle \cup_i}  (X_i,\mathbb{P}_i)$. For instance, in adversarial settings, we consider as a source domain $X={\displaystyle \cup_i}  (X_i,\mathbb{P}_{Attack_i})$, composed of the union of distinct subsets $X_i$ from different attack distributions $\mathbb{P}_{Attack_i}$ (i.e., $Attack_i \neq Attack_j$ if $i \neq j$). We note that depending on the defender's preference and priorities (e.g., strength of attacks), the proportion of each attack distribution may vary. In particular, each attack distribution $Attack_i$ is represented by a subset of adversarial examples in the source domain $X$. For instance, in Section \ref{result:OOD_adv}, we perform our experiments using $X=\{X_{C\&W},X_{FGSM},X_{PGD}\}$ with equal proportions of each attack  distribution ($\frac{1}{3}$ for each attack).

Although this design might lead to a better generalization of the translation capabilities, it might also decrease the performance compared to separate standalone translator trained on each attack as one source distribution (e.g., FGSM2Benign, PGD2Benign, etc). Thus, we further investigate another alternative design that is likely to maintain the performance of each separate translator while also generalizes the translation capabilities to any attack. we call it \textit{Ensemble of Translators}. 

\vspace*{-1em}
\subsection{Ensemble of Translators}\label{subsec:ensemble-trans}
\begin{figure}[t!]
    
    \centering
    
    \includegraphics[width=\columnwidth]{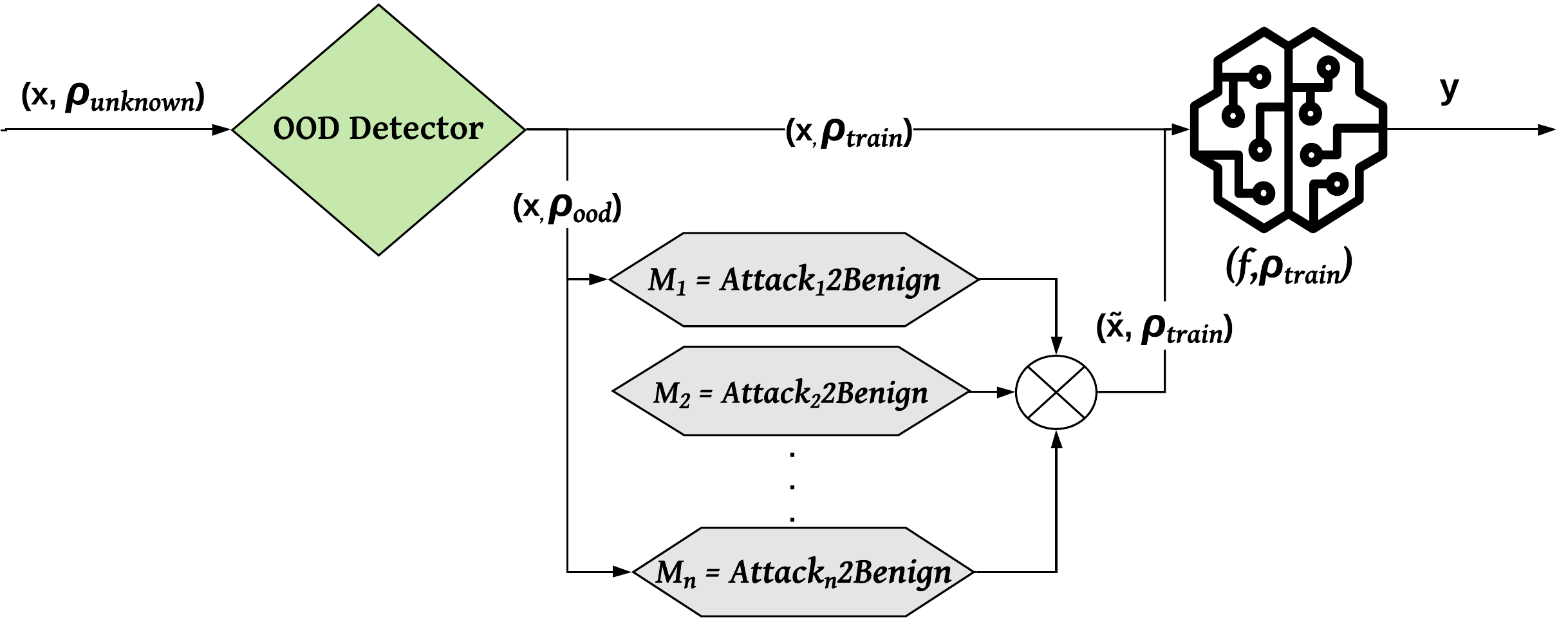}
   \vspace*{-1em}
    \caption{Our approach with ensemble of translators $\mathcal{M} = (\mathcal{M}_1,...,\mathcal{M}_n)$.}
    \vspace*{-1em}
    \label{fig:ens-gan}

\end{figure}

\textbf{Multiple Standalone Translators:} For this design, multiple standalone translators are deployed together. In adversarial settings, each standalone translator is trained on a source domain of only one attack distribution. Consequently, the overall in-distribution mapping model is defined as $\mathcal{M} = (\mathcal{M}_1,...,\mathcal{M}_n)$, where each translator $\mathcal{M}_i$ is trained to translate test samples from the distribution $\mathbb{P}_{Attack_i}$ into the original distribution $\mathbb{P}_{data}$ of the target model $f$. As illustrated in Figure \ref{fig:ens-gan}, given an adversarial query ($x,\mathbb{P}_{adv}$), each module $\mathcal{M}_i$ generates a candidate translation $\Tilde{x}_i$. Hence, a selection approach is needed to pick the best input translation and feed it to $f$ for classification. Two selection approaches standout:


  \noindent \textbf{Majority Vote:}
First, we consider the majority vote approach. Particularly, $f$ predicts the labels of all possible input translations $\{\Tilde{x}_i\}_{i\leq n}$. The label $y_i=f(\Tilde{x_i})$ with the highest number of occurrence across all predictions  is returned as the final prediction.

\noindent  \textbf{Highest Confidence:}
Another method is to pick the predicted label that has the highest confidence of the model $f$, i.e., prediction probability $P_f(y)$. For each possible input $\Tilde{x_i}$, $f$ attributes a prediction probability to each label $y \in Y$. The selected label is the one that has the highest prediction probability, in total across all inputs $\{\Tilde{x}_i\}_{i\leq n}$. Formally,
\begin{equation*}
  y = \max_{y\in Y} \{\sum_{\{\Tilde{x}_i\}_{i\leq n}}    P_{f|\Tilde{x}_i}(y) \}
  \vspace*{-1em}
\end{equation*}


It is noteworthy that the \textit{ensemble translators} design can be extended to, additionally, include other translators trained to translate \textit{non-adversarial} OOD samples ($x,\mathbb{P}_{OOD}$) where $\mathbb{P}_{OOD}\neq\mathbb{P}_{adv}$ (e.g., blur2clear, sharp2normal, etc). Hence, realizing the ultimate goal of OOD generalization, on adversarial and non-adversarial inputs.

In Section \ref{result:OOD_adv}, we evaluate all the discussed designs and compare their effectiveness against unexpected attacks with $\mathbb{P}_{adv}=\mathbb{P}_{unknown}$. Next, we go through the architecture of an in-distribution mapping model with respect to the domain at hand (image classification). 
\vspace*{-1em}
\subsection{In-Distribution Mapping Model Architecture}\label{subsec:translator-archs}

Our in-distribution mapping model is defined to specifically translate OOD test images into in-distribution, with respect to the design alternatives discussed earlier. Thus, we leverage Image-to-Image translation methods as tools to achieve our in-distribution mapping goal. We particularly study two GAN-based methods that currently are benchmarks for Image-to-Image translation applications:  CycleGAN~\cite{CycleGAN2017} and pix2pix~\cite{isola2018imagetoimage}.
\vspace*{-1em}
\subsubsection{M =  CycleGAN}
As described in Section \ref{sec: bground}, Image-to-Image translation approaches have been demonstrated to be successful for many tasks such as translating: summer landscapes to winter landscapes (or the reverse), paintings to photographs (or the reverse), and horses to zebras (or the reverse) \cite{CycleGAN2017}. However, most Image-to-Image translation methods require a dataset comprised of paired examples ~\cite{long2015fully,isola2018imagetoimage}. For instance, a summer2winter landscape translator model needs a large dataset of many examples of input images $X=\{x_i\}_{i\leq d}$ (i.e., summer landscapes) and the same number of images with the desired modification that can be used as expected output images $\Tilde{X}=\{\Tilde{x_i}\}_{i\leq d}$ (i.e., winter landscapes). Gathering paired datasets is challenging for several real-world applications. In CycleGAN~\cite{CycleGAN2017}, however, there is no need to have paired images, which makes it preferred for our approach: it can be trained using source training images $X$ that are unrelated to the target training images $\Tilde{X}$ (i.e., unpaired dataset).

As a defense against adversarial examples, our approach can be deployed with CycleGAN as in-distribution mapping architecture by training a generator $G: (X_{adv}, \mathbb{P}_{ood}) \rightarrow (X_{benign},\mathbb{P}_{data})$ that generates a benign version of an adversarial input. With respect to the cycle-consistency losses defined in Section \ref{subsec:I2I}, a reverse generator $F: (X_{benign},\mathbb{P}_{data}) \rightarrow (X_{adv}, \mathbb{P}_{ood})$ is additionally trained. Since CycleGAN uses unpaired training datasets, the training of $G$ is guided by the training of $F$ in order to ensure that the input image $x$ can be reconstructed using the generated image $G(x)=\Tilde{x}$. The generator $F$ needs to satisfy the cycle-consistency constraint (i.e., $F(G(x)=x$), which mitigates the impact of using unpaired training data (i.e., unsupervised learning).

In Section \ref{sec: eval}, we conduct extensive experiments to evaluate the effectiveness of our approach using $\mathcal{M} = $ CycleGAN to translate OOD adversarial examples to their benign equivalents that lie in the same distribution as $f$'s training set.
\vspace*{-1em}
\subsubsection{M =  pix2pix}
Adversarial examples are generated by performing perturbations on originally benign data. Consequently, in this particular case, a paired dataset can be easily prepared. More precisely, we use the original benign version of adversarially-generated samples as the target distribution to train $\mathcal{M}$. Therefore, we additionally consider supervised Image-to-Image translation methods as options to define $\mathcal{M}$ (i.e., pix2pix).

As illustrated in Section \ref{subsec:I2I}, pix2pix trains a conditional GAN to generate images in the target distribution subject to the condition of the input source distribution. In the original paper \cite{isola2018imagetoimage}, Isola et al. show that a U-Net architecture, which is a convolutional network for image segmentation \cite{ronneberger2015unet}, was particularly suitable for their experiments (e.g., Cityscapes labels2photos, Edges2photo). Segmentation models (such as U-net) are useful especially for images that contain more than one object. Thus, in our experiments, we consider a ResNet generator instead, given that ResNet models have been particularly successful on ImageNet and CIFAR10 \cite{he2015deep}.

\begin{figure}[t!]
    
    \centering
    \scalebox{0.95}{
    \includegraphics[width=\columnwidth]{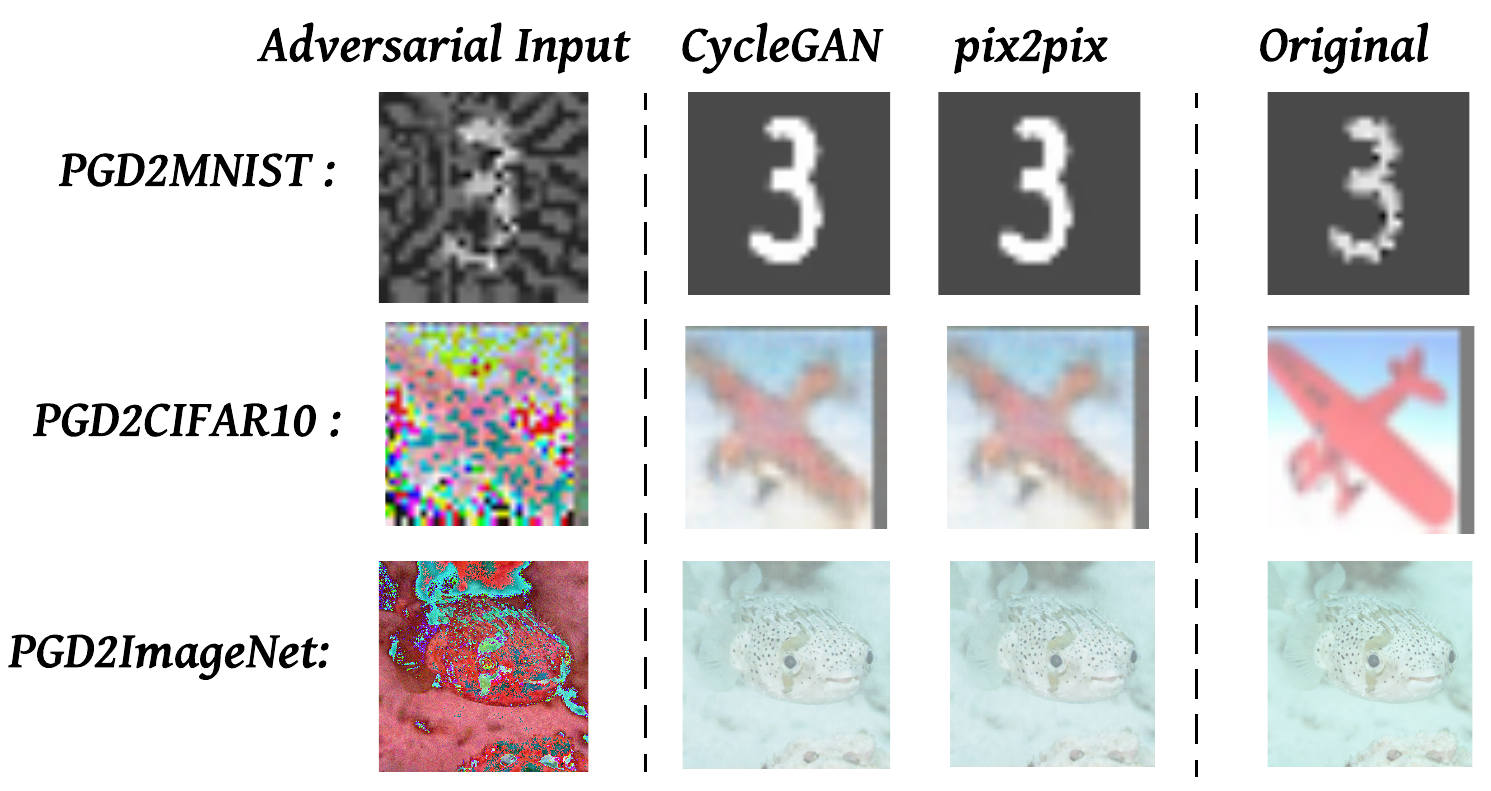}}
   \vspace*{-1em}
    \caption{An illustration of Image-to-Image translation of OOD PGD-adversarial examples on normalized samples, to in-distribution benign examples using CycleGAN and pix2pix on MNIST, CIFAR10, and ImageNet images.}
    \vspace*{-1.75em}
    \label{fig:examples}

\end{figure}

Figure \ref{fig:examples} shows three examples of translating adversarial images from the PGD attack distribution (on normalized data) to benign images close to the training distribution in three datasets (MNIST, CIFAR10, and ImageNet) using $\mathcal{M} =$ CycleGAN = PGD2Benign, and $\mathcal{M} =$ pix2pix = PGD2Benign. For all three datasets, PGD training samples are used as the source distribution $\mathbb{P}_{PGD}$ to train $\mathcal{M}: (X,\mathbb{P}_{PGD}) \rightarrow (\Tilde{X},\mathbb{P}_{benign})$. Thus, $\mathcal{M}$ can successfully map an OOD input (PGD image in this case) to the distribution of benign data.

\section{Evaluation}\label{sec: eval}
We evaluate our approach on three benchmark image classification datasets across alternative designs of $\mathcal{M}$. 
First, we describe the experimental setup (Section \ref{sec:setup}). Second, we interpret our results in two OOD problem settings: against adversarial OOD examples (Section \ref{result:OOD_adv}) and against benign OOD examples (Section \ref{result:OOD_benign}). 
\vspace*{-1em}
\subsection{Experimental Setup}\label{sec:setup}

\subsubsection{Datasets and Models}
\textbf{Datasets:} We use three benchmark image classification datasets of different scales (i.e., MNIST~\cite{MNIST}, CIFAR10~\cite{cifar}, and ImageNet~\cite{ImageNet}). First, we reach a proof-of-concept on MNIST. Second, we evaluate the effectiveness of our approach on CIFAR10 as a more realistic dataset. Finally, we use ImageNet to evaluate to what extent our approach scales to larger datasets. \textbf{MNIST} is hand-written digits classification dataset that contains 60K training samples of gray-scale images of the size 28x28 and 10K test samples. \textbf{CIFAR10} consists of 60K 32x32 colour images in 10 classes, with 6K images per class (e.g., airplane, cat, dog, etc). 50K samples are used for training and 10K for testing. In regards to \textbf{ImageNet}, we specifically consider the widely-used data of the Large Scale Visual Recognition Challenge (ILSVRC 2012-2017)~\cite{ILSVRC15}. It spans 1K classes and contains 1,281,167 training images, 50K validation images and 100K test images.

\textbf{Models:} On MNIST, we train a CNN model that reaches a test accuracy of $98.96\%$ (on 10K samples). For CIFAR10, we adopt a SimpleDLA network that reaches one of the best performances compared to the state-of-the-art accuracy of $95.19\%$ originally proposed in \cite{CIFAR10-model}. For ImageNet, we use a pretrained ResNet50 model available via PyTorch~\cite{8100117}. On the first 100 classes of the validation data (i.e., 5K samples), it reaches an accuracy of $83.58\%$. Details of each model's architecture are included within our publicly available code\footnote{\url{https://github.com/OOD2IID/code}}.
\vspace*{-1em}
\subsubsection{Image-to-Image Translator Models}
\textbf{Architectures:} As described in Section \ref{subsec:translator-archs}, we leverage \textit{CycleGAN} for unpaired datasets and \textit{pix2pix} for paired datasets. For CycleGAN, inline with the original implementation~\cite{CycleGAN2017}, we use 9 blocks of ResNet models as generators $G$ and $F$ and a PatchGAN classifier as discriminators (i.e., $D_G$ and $D_F$). It can classify whether 70×70 overlapping patches are real or fake. Such a patch-level discriminator architecture has fewer parameters than a full-image discriminator and works well on arbitrarily-sized images in a fully convolutional fashion \cite{CycleGAN2017}. As for pix2pix, we use 6 blocks of ResNet models for the generator $G$ and PatchGAN classifier as discriminator. For MNIST, all models are trained on 500 images from the source distribution and their equivalent in the target distribution. As for CIFAR10 and ImageNet, we use 1000 training images in each distribution (i.e., source and target distributions). In order to reduce the performance overhead of Image-to-Image translation models, we consider only the first 100 of the 1000 classes of ImageNet to train and test $\mathcal{M}$.   

\textbf{Source Distributions:}
We evaluate our approach in two complementary settings of the OOD problem: (a) when the OOD test input is adversarial, i.e., intentionally crafted by an adversary (Section \ref{result:OOD_adv}) and (b) when the test input is OOD due to natural distribution shift (Section \ref{result:OOD_benign}). We note that a benign input can be OOD due to factors such as the level of brightness, image sharpness, camera resolution, contrast, etc. 

For setting (a), we test all the definitions of $\mathcal{M}$ discussed in Sections \ref{subsec:standalone-trans} and \ref{subsec:ensemble-trans} which include a standalone translator with a single source distribution (PGD, FGSM or SPSA), a standalone translator with a mixture of source distributions (FGSM, PGD, and C\&W ), and an ensemble of translators $\mathcal{M}=\{\mathcal{M}_1 = PGD2Benign, \mathcal{M}_2 = FGSM2Benign, \mathcal{M}_3=SPSA2Benign \}$.
For setting (b), we conduct two experiments where $\mathcal{M}$ is a standalone translator. To train the translator $\mathcal{M}$, in the first experiment we consider dark images as the source distribution and in the second experiment, we use images with high sharpness as OOD source distribution. More details about each experiment are provided in sections \ref{result:OOD_adv} and \ref{result:OOD_benign} as we discuss the results. 

\textbf{Target Distribution:}
Inline with the purpose of our approach, the target distribution is always the original training distribution $\mathbb{P}_{train}$ used to train the model $f$. It is one of MNIST, CIFAR10, or ImageNet training distributions. 

\textbf{Notations:} In all the next figures and tables, we use the abbreviated notation "mix" to refer to the source distribution of a mixture of attacks, "Ens-$\mathcal{M}$-MV" to denote ensemble translators with majority vote and "Ens-$\mathcal{M}$-HC" to denote ensemble translators with highest confidence prediction.
\vspace*{-1em}
\subsubsection{Evaluation Metrics}
Our evaluation relies on two complementary metrics, prediction \textit{Accuracy} and \textit{Relative Robustness}.

\textbf{Accuracy:} It is used to compute the rate of correct predictions out of the total number of test samples. We use this metric to compare the model's prediction accuracy on OOD data before and after using our approach (Section \ref{result:OOD_benign}).

\textbf{Relative Robustness (RR):} The robustness of a ML model on adversarial data is relative to its performance on benign data. For instance, the best neural network trained on CIFAR10 can only reach a maximum of $~95.19\%$ accuracy on benign data (i.e., error of $5\%$)~\cite{CIFAR10-model}. In order to ensure a fair evaluation of robustness, we evaluate the model's performance under attack, relative to its performance before attack. In particular, we compute a metric that compares the number of correct predictions made by $f$ under attack with the number of correct predictions on benign data. We call this metric the Relative Robustness (RR). Formally, it is defined as:
\begin{equation}
  RR (\%) = \frac{\sum_{x\in X} \{f(x+\delta)=y_{true}\}}{\sum_{x\in X} \{f(x)=y_{true}\}} \times 100
  \label{eq:RR}    
\end{equation}
where $x\in X$ is the test sample, $y_{true}$ is the true label of $x$, and  $\delta$ is the attack's perturbation.

The RR metric is utilized in Section \ref{result:OOD_adv} to evaluate the effectiveness of our approach in improving robustness of $f$. With respect to Equation \ref{eq:RR}, on benign data ($\delta=0$), $RR=100\%$. On adversarial data, if $RR$ is close to $100\%$, then the accuracy on adversarial data is close to the accuracy on benign data which reflects high robustness. Otherwise, the model is less robust.  The case of $RR>100\%$ is technically possible, however, it is unlikely to occur since it means that the model is more accurate on adversarial data than on benign data.

\subsubsection{Benchmark Attacks}
\vspace*{-1em}
We consider a range of benchmark evasion attacks for our experiments. In particular, we use three white-box attacks (i.e., FGSM~\cite{FGSM}, PGD~\cite{PGSM}, and C\&W~\cite{CW}) and a black-box attack (i.e., SPSA~\cite{SPSA}) to cover complementary threat models and diverse attack strengths. In a white-box setting, we use FGSM as a fast one-step attack with low execution overhead. Additionally, we use PGD as an improved gradient-based attack that leads to a higher evasion rate, with reasonable performance overhead. Finally, we consider the C\&W attack, one of the strongest evasion attacks despite its high performance overhead. In a black-box setting, we use SPSA as one of the representative attacks that performs perturbations without any access to the model's architecture or dataset.

All attacks are subject to $||\delta||<\epsilon$, where $\epsilon$ is the maximum perturbation size. For each dataset, the maximum perturbation size $\epsilon$ is picked such that the produced adversarial image is still recognizable by the human eye. To that end, we select $\epsilon = 0.3$ for MNIST and $\epsilon = 0.2$ for CIFAR10. For ImageNet, our approach maintains its robustness for $\epsilon=0.2$. However, we choose to follow the state-of-the-art by using a maximum of $\epsilon=8/255 \approx 0.031$ so as to conduct a fair comparison with prior defenses \cite{salman2020adversarially}. For all attacks, we use $||.||_\infty$ norm except C\&W which supports $||.||_2$.
\vspace*{-1.5em}
\subsubsection{Benchmark Defenses}
\vspace*{-1em}
So far, the closest the state-of-the-art got to OOD generalization for adversarial robustness is through \textit{adversarial training}. An adversarially-trained model is in fact an improved model that generalizes to OOD adversarial examples, which makes it more robust, albeit its penalty on benign accuracy. As a result, we consider adversarial training techniques as benchmark defenses to compare our approach with the state-of-the-art.

\begin{figure*}[t!]
    
    \centering
    \includegraphics[width=.95\textwidth]{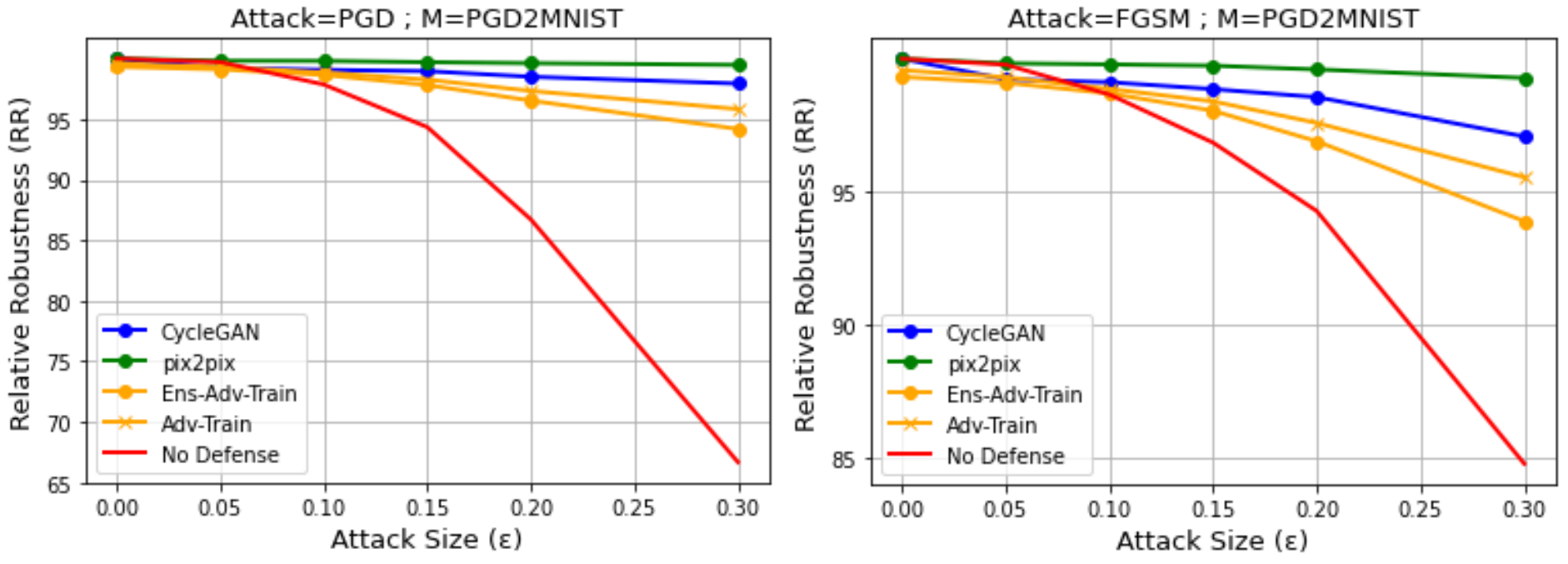}
    \vspace*{-1em}
    \caption{PGD (left) and FGSM (right) against our approach with $\mathcal{M}$=PGD2MNIST and adversarial training techniques.}
    \label{fig:pgd2mnist}
    \vspace*{-1em}
\end{figure*}

\begin{figure*}[t!]
    
    \centering
    \includegraphics[width=.95\textwidth]{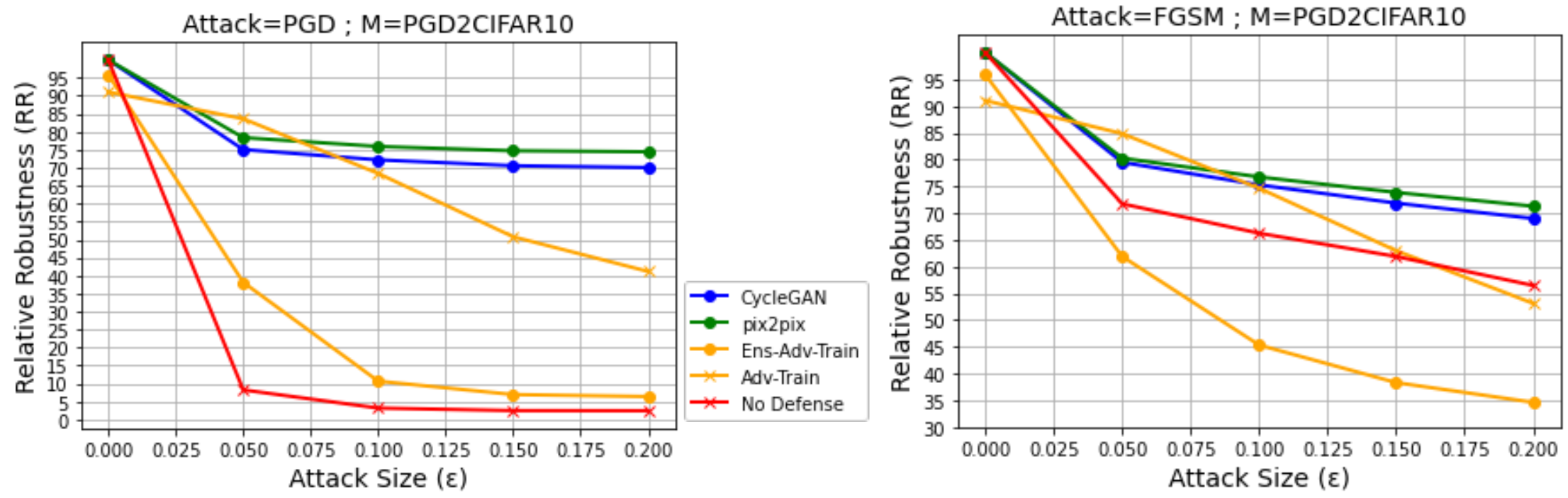}
    \vspace*{-1em}
    \caption{PGD (left) and FGSM (right) against our approach with $\mathcal{M}$=PGD2CIFAR10 and adversarial training techniques.}
    \label{fig:pgd2cifar10}
     \vspace*{-1em}
\end{figure*}

\begin{figure*}[t!]
    
    \centering
    \includegraphics[width=.95\textwidth]{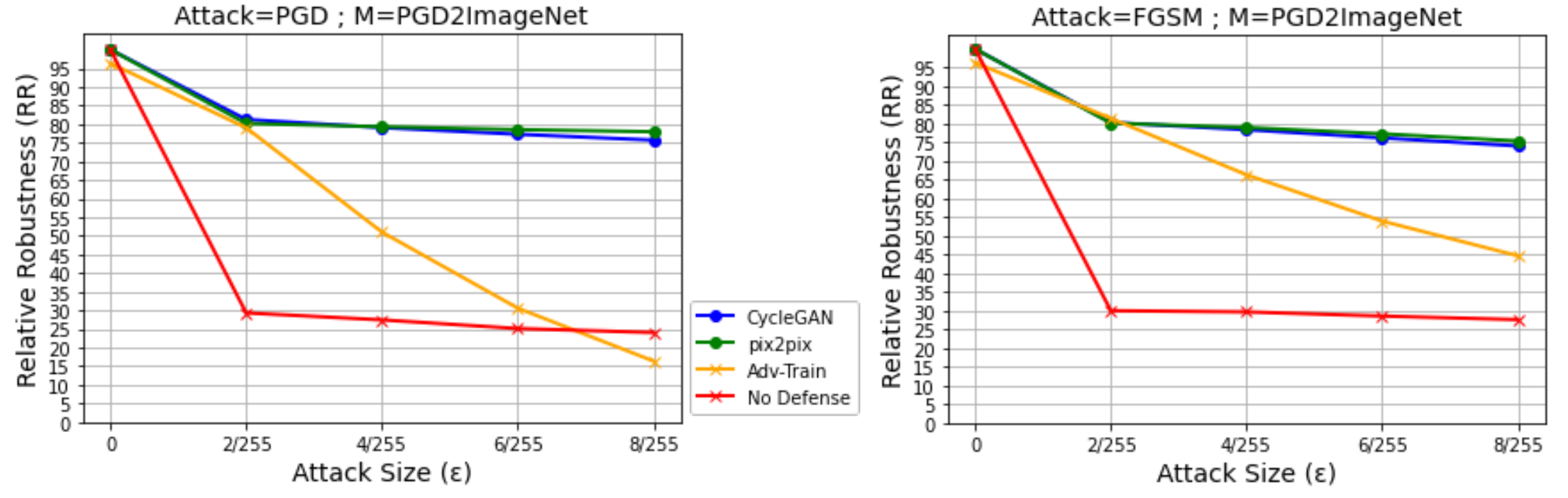}
   \vspace*{-1em}
    \caption{PGD (left) and FGSM (right) against our approach with $\mathcal{M}$=PGD2ImageNet and adversarial training techniques.}
    \label{fig:pgd2imagenet}
     \vspace*{-1em}
\end{figure*}

\textbf{Adversarial training:}  A training~\cite{salman2020adversarially,kurakin2017adversarial} scheme of exposing the model $f$ to adversarial examples during the training phase in order to learn the correct labels of adversarial test samples. It has been considered as one of the most effective and practical defenses against evasion attacks. However, its main limitation is that it sacrifices the model’s accuracy on clean data. As we show in our results in Section \ref{result:OOD_adv} we overcome this limitation by preserving accuracy on benign data while significantly improving robustness to OOD inputs. 

\textbf{Ensemble Adversarial training:} Another technique of adversarial training is to train the model on adversarial examples that are generated using a set of different models as target models. This technique is more effective against multiple-step attacks, especially black-box attacks \cite{EnsembelAdvTrain18}.

\textbf{Notations:} In all subsequent figures, tables, and discussion we use ``Adv-Train" for adversarial training and ``Ens-Adv-Train" for ensemble adversarial training.

\subsection{Generalization to OOD Adversarial Examples}
\label{result:OOD_adv}
First, we perform a comparative experiment of our approach against adversarial training techniques. Then, we compare the performances of different designs of $\mathcal{M}$ with respect to the options discussed in Sections \ref{subsec:standalone-trans} and \ref{subsec:ensemble-trans}.
\vspace*{-1em}
\subsubsection{Comparison with Benchmark Defenses}

For this experiment, we consider the case where the defender deploys a standalone distribution translator trained to translate PGD adversarial data into in-distribution data (e.g., $\mathcal{M}$=PGD2MNIST,  $\mathcal{M}$=PGD2CIFAR10, and $\mathcal{M}$=PGD2ImageNet). For each dataset,  we use both CycleGAN and pix2pix. To ensure a fair comparison with the benchmark adversarial training techniques, we similarly use PGD attack to perform Adv-Train and ens-adv-train on the target model $f$. Figures \ref{fig:pgd2mnist}, \ref{fig:pgd2cifar10}, and \ref{fig:pgd2imagenet} show the RR values using our OOD generalization-based defense compared to adversarial training techniques against PGD (left) and FGSM (right) attacks, respectively, for MNIST, CIFAR10, and ImageNet. We additionally refer to Table \ref{tab:summary} to examine the RR values of different defenses against C\&W and SPSA.

\textbf{Robustness against PGD and FGSM:} Figure \ref{fig:pgd2mnist} shows that, on average, unlike Adv-Train ($RR \approx 95\%$) and Ens-Adv-Train ($RR\approx93\%$), our approach with $\mathcal{M}$=PGD2MNIST maintains RR $>98\%$ on PGD and FGSM attacks, even for higher attack size ($\epsilon=0.3$). These results suggest that using our approach, the accuracy of $f$ under attack is almost the same as its accuracy on benign data. Comparable findings are observed on CIFAR10 and ImageNet (Figures \ref{fig:pgd2cifar10} and \ref{fig:pgd2imagenet}). In particular, with $\mathcal{M}$=PGD2CIFAR10 and $\mathcal{M}$=PGD2ImageNet, our approach outperforms adversarial training techniques by a significant margin. More precisely, considering the worst case scenario of both attacks (i.e., CIFAR10: $\epsilon=0.2$, ImageNet:  $\epsilon=8/255\approx 0.031$ ), our approach outperforms Adv-Train and Ens-Adv-Train, respectively by a margin of $25.72\%$ and $49.19\%$ on CIFAR10, averaged over FGSM and PGD. It reaches RR $\approx 75\%$ (Figure \ref{fig:pgd2cifar10}). Additionally, it beats Adv-Train on ImageNet by a margin of $46.14\%$ averaged over FGSM and PGD to reach  $RR=77.96\%$ against PGD and $RR= 75.44\%$ against FGSM (Figure \ref{fig:pgd2imagenet} and Table \ref{tab:summary}). We note that for ImageNet we consider only Adv-Train as benchmark defense due to high performance overhead of adversarial training techniques (especially Ens-Adv-Train).

\noindent \fbox{\parbox{.96\linewidth}{\textbf{Conclusion 1:} Our approach generalizes to OOD adversarial examples far better than state-of-the-art defenses.}}

\begin{table*}[t]

\centering

\scalebox{.95}{
\begin{tabular}{!{\color{black}\vrule}l!{\color{black}\vrule}l!{\color{black}\vrule}l!{\color{black}\vrule}l!{\color{black}\vrule}l!{\color{black}\vrule}l!{\color{black}\vrule}l!{\color{black}\vrule}l!{\color{black}\vrule}l!{\color{black}\vrule}l!{\color{black}\vrule}l!{\color{black}\vrule}l!{\color{black}\vrule}l!{\color{black}\vrule}!{\color{black}\vrule}l!{\color{black}\vrule}l!{\color{black}\vrule}l!{\color{black}\vrule}l!{\color{black}\vrule}l!{\color{black}\vrule}l!{\color{black}\vrule}l!{\color{black}\vrule}l!{\color{black}\vrule}l!{\color{black}\vrule}l!{\color{black}\vrule}l!{\color{black}\vrule}l!{\color{black}\vrule}l!{\color{black}\vrule}!{\color{black}\vrule}l!{\color{black}\vrule}l!{\color{black}\vrule}l!{\color{black}\vrule}l!{\color{black}\vrule}l!{\color{black}\vrule}l!{\color{black}\vrule}l!{\color{black}\vrule}l!{\color{black}\vrule}l!{\color{black}\vrule}l!{\color{black}\vrule}l!{\color{black}\vrule}l!{\color{black}\vrule}l!{\color{black}\vrule}!{\color{black}\vrule}l!{\color{black}\vrule}l!{\color{black}\vrule}l!{\color{black}\vrule}l!{\color{black}\vrule}l!{\color{black}\vrule}l!{\color{black}\vrule}l!{\color{black}\vrule}l!{\color{black}\vrule}l!{\color{black}\vrule}l!{\color{black}\vrule}l!{\color{black}\vrule}l!{\color{black}\vrule}l!{\color{black}\vrule}} 
\hline
 \multicolumn{3}{|c!{\color{black}\vrule}}{}&\textbf{No Attack}& \textbf{FGSM} & \textbf{PGD} & \textbf{C\&W} & \textbf{SPSA}\\
\hline
\scriptsize1& \multirow{3}{*}{\bf{No Defense}} & MNIST& 100\% &84.79\% &66.69\% &0.07\%& 85.13\%\\
\cline{3-8}
\scriptsize2&& CIFAR10 &100\%&56.48\%&2.41\%&0.01\%&47.70\%\\
\cline{3-8}
\scriptsize3&& ImageNet &100\%&27.63\%&24.06\%&0.0\%&65.79\%\\
\hline
\hline
\scriptsize4& \multirow{2}{*}{\textbf{Ens-Adv-Train}} & MNIST& 99.32\% &93.90\% &94.19\% &1.99\%& 94.14\%\\
\cline{3-8}
\scriptsize5&& CIFAR10 &95.87\%&34.63\%&6.35\%&0.71\%&41.11\%\\
\hline
\scriptsize6& \multirow{3}{*}{\textbf{Adv-Train}} & MNIST& 99.57\% &95.55\% &95.83\% &0.86\%&95.59\% \\
\cline{3-8}
\scriptsize7&& CIFAR10 &91.06\%&53.13\%&41.15\%&0.52\%&68.41\%\\
\cline{3-8}
\scriptsize8&& ImageNet &96.33\%&44.69\%&16.39\%&0.0\%&82.01\%\\
\hline
\hline
\scriptsize9& \multirow{3}{*}{\textbf{Our approach with CycleGAN}}  & $\mathcal{M}$ = PGD2MNIST &100\%&97.08\%&97.94\%&98.20\%&97.44\%\\
\cline{3-8}
\scriptsize10&& $\mathcal{M}$ = PGD2CIFAR10 &100\%&69.00\%&70.05\%&48.90\%&64.67\%\\
\cline{3-8}
\scriptsize11&& $\mathcal{M}$ = PGD2ImageNet &100\%&74.06\%&75.71\%&\color{brown}{\textbf{46.03\%}}&80.27\%\\
\hline
\hline
\scriptsize12& \multirow{8}{*}{\textbf{Our approach with Pix2Pix}}  & $\mathcal{M}$ = PGD2MNIST &100\%&\color{blue}\textbf{99.28\%}&\color{blue}\textbf{99.48\%}&\color{blue}\textbf{99.12}\%&\color{blue}\textbf{99.27\%}\\
\cline{3-8}
\scriptsize13&& $\mathcal{M}$ = PGD2CIFAR10 &100\%&71.29\%&74.43\%&49.50\%&67.58\%\\
\scriptsize14&& $\mathcal{M}$ = SPSA2CIFAR10 &100\%&71.26\%&74.37\%&45.58\%&72.69\%\\
\scriptsize15&& $\mathcal{M}$ = FGSM2CIFAR10 &100\%&\color{green}{\textbf{74.96}}\%&\color{green}{\textbf{78.95\%}}&47.44\%&\color{green}{\textbf{72.70\%}}\\


\scriptsize16&& $\mathcal{M}$ = MIX2CIFAR10 &100\%&60.41\%&60.51\%&\color{green}{\textbf{68.69\%}}&56.54\%\\
\scriptsize17&& $\mathcal{M}$ = Ens-$\mathcal{M}$-MV (CIFAR10) &100\%&72.72\%&76.48\%&48.44\%&70.10\%\\
\scriptsize18&& $\mathcal{M}$ = Ens-$\mathcal{M}$-HC (CIFAR10) &100\%&73.53\%&77.33\%&49.81\%&70.38\%\\
\cline{3-8}
\scriptsize19&& $\mathcal{M}$ = PGD2ImageNet &100\%&\textcolor{brown}{\textbf{75.40\%}}&\textcolor{brown}{\textbf{77.96\%}}&25.40\%&\textcolor{brown}{\textbf{83.10\%}}\\
\hline

\end{tabular}

}


\fbox{\begin{tabular}{lccc}
&Best result per attack on:\\
\textcolor{blue}{$\blacksquare$}  MNIST  &
\textcolor{green}{$\blacksquare$}  CIFAR10 &
\textcolor{brown}{$\blacksquare$} ImageNet 
\end{tabular}}

\vspace*{-1em}
\caption{Summary of the Relative Robustness (RR) of different definitions of $\mathcal{M}$ and adversarial training (Adv-Train) and ensemble adversarial training (Ens-Adv-Train) for MNIST, CIFAR10, and ImageNet. Attack\_size $\epsilon = 0.3$ for MNIST, $0.2$ for CIFAR10, and $8/255\approx 0.031$ for ImageNet. All attacks are performed with $||.||_\infty$ except C\&W, which is performed with $||.||_2$.}
\label{tab:summary}
\vspace*{-1em}
\end{table*}

\textbf{Accuracy preservation on benign data:} On top of its convincing robustness over benchmark defenses, Figures \ref{fig:pgd2mnist}, \ref{fig:pgd2cifar10}, and \ref{fig:pgd2imagenet} show that our defense does not sacrifice accuracy on benign data ($RR=100\%$ for $\epsilon=0$), unlike adversarial training techniques that sacrifice the accuracy on benign data by an average margin (over Adv-Train and Ens-Adv-Train) of $\approx 1\%$ on MNIST, $6.53\%$ on CIFAR10, and $3.67\%$ on ImageNet. This preservation of benign accuracy is attributed to the OOD detector that successfully differentiates between adversarial and benign test data, and feeds benign (in-distribution) data directly to $f$ without involving $\mathcal{M}$.

\noindent \fbox{\parbox{.96\linewidth}{\textbf{Conclusion 2:} Unlike state-of-the-art defenses, our approach generalizes to OOD adversarial examples without penalizing accuracy on benign test data.}}

\textbf{CycleGAN vs. pix2pix:} Although both Image-to-Image translation architectures allow our approach to stand against adversarial example attacks and surpass the state-of-the-art results, we note that, on all three datasets, pix2pix architecture is more effective than CycleGAN in the translation of OOD examples to in-distribution equivalents. More precisely, pix2pix outperforms CycleGAN by an average margin of $\approx2\%$, $\approx5\%$,  and $\approx3\%$, respectively, on MNIST, CIFAR10, and ImageNet, over both attacks (i.e., FGSM and PGD). We recall that pix2pix is a supervised learning approach, while CycleGAN is unsupervised (i.e., relies on cycle consistency) which explains the observed advantage of pix2pix.

\noindent \fbox{\parbox{.96\linewidth}{\textbf{Conclusion 3:} pix2pix offers better OOD to in-distribution translation success than CycleGAN on all three datasets.}}

\textbf{Robustness against unexpected attacks (C\&W and SPSA):} We recall that Adv-Train and Ens-Adv-Train are performed using PGD-generated training samples. Additionally, we recall that, for a fair comparison with our approach, we, similarly, consider only translators trained on PGD samples for this experiment. Using attacks that were not incorporated in the training of $\mathcal{M}$, we now evaluate the robustness of our approach against C\&W and SPSA as ``unexpected'' attacks.

From Table \ref{tab:summary},  we focus only on results for $\mathcal{M}$ = PGD2MNIST (rows 9 and 12), $\mathcal{M}$ = PGD2CIFAR10 (rows 10 and 13), $\mathcal{M}$ = PGD2ImageNet (rows 11 and 19). Compared to Adv-Train and/or Ens-Adv-Train results (rows 4-8) over all three datasets, we confirm the previous results on different attacker (i.e., C\&W and SPSA). More precisely, while adversarial training techniques are basically defeated by the C\&W attack (i.e., $RR\approx1\%$), our approach delivers a robust model on MNIST (row 12: $RR = 99.12\%$) and a significant improvement on CIFAR10 (row 13: $RR= 49.50\%$) and ImageNet (row 11: $RR = 46.03\%$) against C\&W. Similar results are observed against a black-box attack (SPSA) , as our defense (MNIST: $RR = 99.27\%$ in row 12, CIFAR10: $RR = 67.58\%$ in row 13, and ImageNet:  $RR = 83.10\%$ in row 19) outperforms Adv-Train (MNIST: $RR=95.59\%$, CIFAR10: $RR = 68.41\%$, ImageNet: $RR = 82.01\%$) and Ens-Adv-Train (MNIST: $RR = 94.14\%$, CIFAR10: $RR = 41.11\%$). Finally, it is noteworthy that, although $\mathcal{M}$ is only trained to translate adversarial images generated using PGD attack subject to $||.||_\infty$, it is able to also translate FGSM and SPSA images. Furthermore, it is able to maintain the same translation performance on C\&W test images even though they are generated subject to $||.||_2$ perturbations, especially on MNIST.
\noindent \fbox{\parbox{.96\linewidth}{\textbf{Conclusion 4:} On MNIST, a standalone translator trained with PGD attack distribution is sufficient to successfully generalize to other attacks such as FGSM, SPSA and C\&W.}}
We note, however, that these findings are not consistent across the three datasets. In particular, again from Table \ref{tab:summary}, using a standalone translator trained only on PGD (row 13), our approach performs less against other attacks, especially on CIFAR10 (i.e., SPSA: $\approx 6.85\%$ less and C\&W: $\approx 24.93\%$ less) compared to its performance against PGD. Although such performance gap is not as large as the one observed for adversarial training techniques (C\&W $\approx 41\%$ less than their performances against PGD), we explore other definitions of $\mathcal{M}$ that can narrow the gap (Section \ref{sec:diff-M}).

\begin{figure*}[t!]
    
    \centering
    \includegraphics[width=\textwidth]{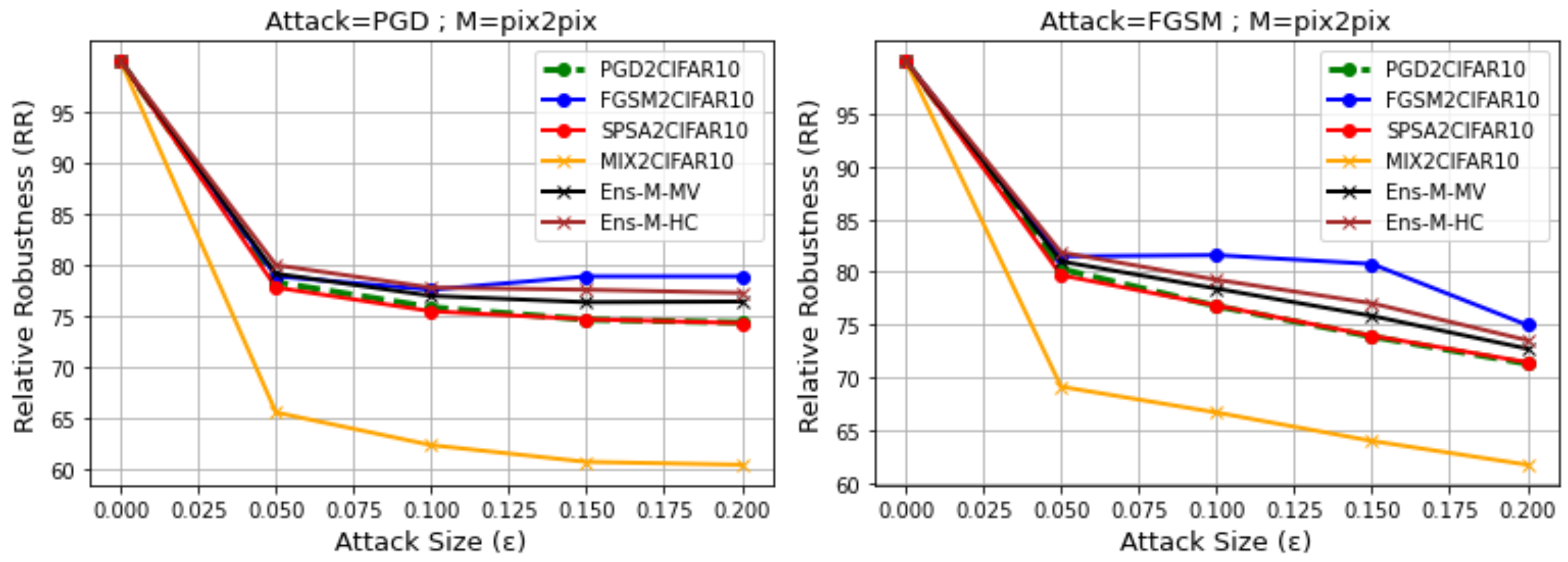}
   \vspace*{-2em}

    \caption{PGD (left) and FGSM (right) against alternative designs of $\mathcal{M}$.}
    \label{fig:M-def}
    \vspace*{-1em}
\end{figure*}

\textbf{Performance Overhead:} Compared to adversarial training techniques, our approach incurrs insignificant performance overhead. We only need to train an OOD detector and an Image-to-Image translator on a small set of images (i.e., MNIST: $500$, CIFAR10: $1000$, and ImageNet: $1000$), while no retraining is needed on the target model $f$. On the other hand, Adv-Train and Ens-Adv-Train require a retraining phase on the model itself using much larger datasets (MNIST: $60$K, CIFAR10: $50$K, and ImageNet:  $>1$M) which leads to considerable delay in the defense deployment. This is especially true for ImageNet due to its complexity and large scale. For instance, using 1 NVIDIA Tesla A100 GPU, $\mathcal{M}$=PGD2ImageNet takes about 25 hours to train for 200 epochs, while Adv-Train on ImageNet takes 10+ days to train for 90 epochs using the same hardware specifications.

\subsubsection{Comparison Between Different Designs of $\mathcal{M}$}\label{sec:diff-M}
We now evaluate alternative designs of $\mathcal{M}$ with respect to the options discussed in Sections \ref{subsec:standalone-trans} and \ref{subsec:ensemble-trans}. At this point, we already established that pix2pix leads to better results than CycleGAN. Hence, for the next experiment, we focus only on $\mathcal{M}$ = pix2pix. Additionally, we exclude MNIST, since $\mathcal{M}$ = PGD2MNIST is sufficient to reach more than $99\%$ robustness on MNIST. Specifically, we focus on CIFAR10.

Figure \ref{fig:M-def} shows the RR results of $\mathcal{M}$= PGD2CIFAR10, FGSM2CIFAR10, SPSA2CIFAR10, and MIX2CIFAR10 as different combinations of the source distribution for standalone translator, and Ens-$\mathcal{M}$-HC, Ens-$\mathcal{M}$-MV as ensemble translators using, respectively, highest-confidence (HC) or majority vote (MV) prediction. We recall that MIX2CIFAR10 is trained on the union of three distributions of attacks (i.e., FGSM, PGD, and C\&W) such that each attack distribution represents $\frac{1}{3}$ of the training set of $\mathcal{M}$. For the ensemble translators method we use an ensemble of three translators: PGD2CIFAR10, FGSM2CIFAR10, and SPSA2CIFAR10.

\textbf{Robustness against PGD and FGS:} From Figure \ref{fig:M-def}, compared to PGD2CIFAR10 (green line) most other translators, including $\mathcal{M}$ = Ens-$\mathcal{M}$, improve the robustness against both attacks (i.e., PGD and FGS) by a margin of $\approx 4\%$, especially for higher attack size $\epsilon>0.05$. This result is true except for $\mathcal{M}=$MIX2CIFAR10 where RR drops to $\approx 61\%$ compared to PGD2CIFAR10 ($RR=74.43\%$ against PGD and $RR=71.29\%$ against FGSM). We note that, using the highest confidence (Ens-$\mathcal{M}$-HC) prediction leads to slightly better results than using majority vote (Ens-$\mathcal{M}$-MV), against gradient-based attacks. Additionally, it is noteworthy that using $\mathbb{P}_{FGSM}$ as source distribution to train a translator $\mathcal{M}$ is the best design to defend against gradient-based attacks on CIFAR10. 

\textbf{Robustness against CW and SPSA (Table \ref{tab:summary}):} Against C\&W, $\mathcal{M}=$MIX2CIFAR10 (row 16) reaches the best robustness $RR=68.69\%$ (row 16) compared to  an average RR $\approx 48\%$ over all other designs (PGD2CIFAR10 in row 13, FGSM2CIFAR10 in row 15, SPSA2CIFAR10 in row 14, and Ens-$\mathcal{M}$ translators in rows 17-18). This observation is explained by including C\&W samples to train MIX2CIFAR10 translator. Both designs of Ens-$\mathcal{M}$ do not improve the robustness against C\&W compared to standalone translator designs, since only SPSA2CIFAR10, FGSM2CIFAR10, and PGD2CIFAR10 translators are adopted for Ens-$\mathcal{M}$ designs (note: no C\&W2CIFAR10 translator is considered). Against SPSA, all designs reach more or less similar results ($\approx 71\%$) on average, except MIX2CIFAR10 which shows $\approx -15\%$ drop in robustness. 


\noindent \fbox{\parbox{.96\linewidth}{\textbf{Conclusion 5:} Among alternative designs of $\mathcal{M}$, FGSM2CIFAR10 is the most performant standalone design overall, while ensemble translators also lead to better robustness to OOD examples with enough number of translators and diverse attack distributions.}}

\vspace*{-1em}
\subsection{Generalization to OOD Benign Data}\label{result:OOD_benign}

We now shift our focus to evaluating our approach for generalization to benign OOD samples so as to account for a non-adversarial setting. We consider two scenarios of test data distribution shift that typically happen due to non-adversarial factors. To that end, we perform two experiments, one on CIFAR10 and the other on ImageNet. On CIFAR10, the distribution shift is caused by changes in \textit{brightness} of test data, while on ImageNet changes in  \textit{sharpness} of test data is the factor. For both experiments, we use pix2pix as image-to-image translator (i.e., for CIFAR10: dark2clear, for ImageNet: sharp2normal). Each translator is trained using 1000 images.

\begin{figure}[t!]
    
    \centering
    \includegraphics[width=\columnwidth]{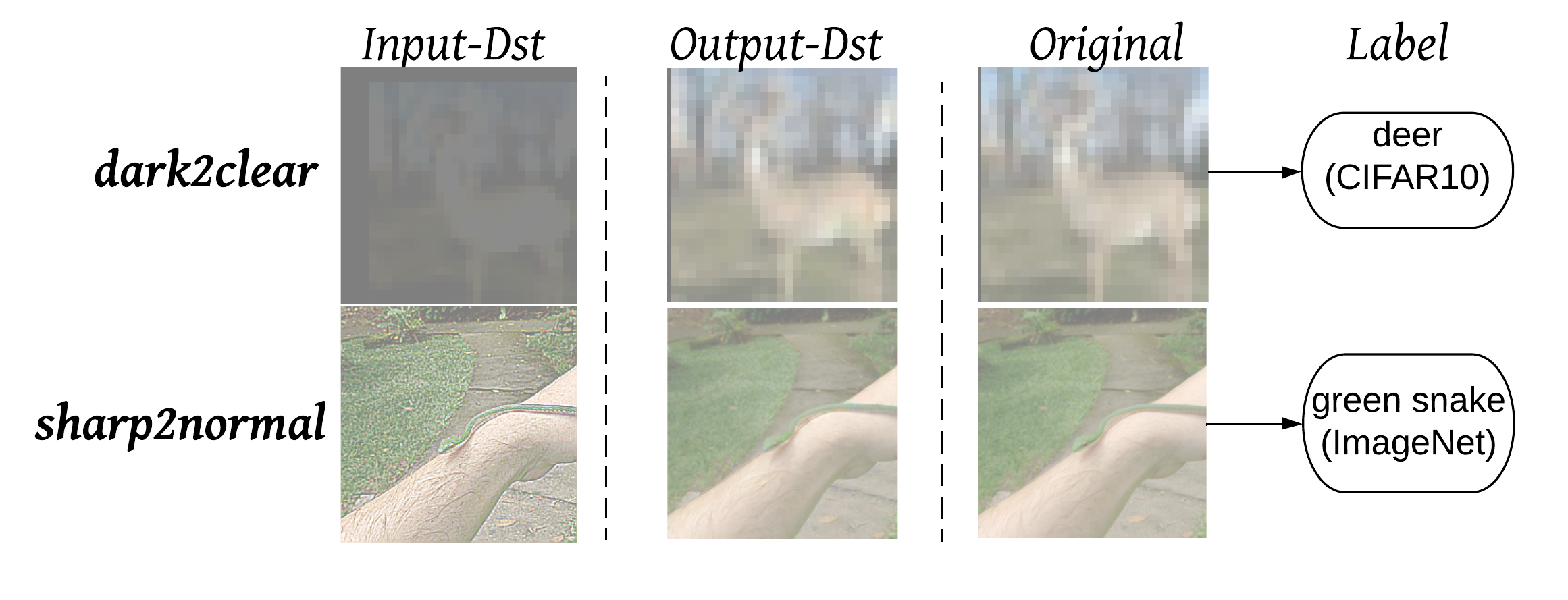}
   \vspace*{-2em}
    \caption{Image-to-Image translation of benign OOD examples
to their in-distribution equivalents using pix2pix on CIFAR10 (dark2clear) and ImageNet (sharp2normal).}
    \label{fig:benign_OOD}
      \vspace*{-1em}
\end{figure}

\textbf{Experiment 1: OOD input due to lower brightness:}
Table \ref{tab:benign_OOD} shows that the accuracy of CIFAR10 model drops from $95.19\%$ to $50.33\%$ on OOD dark images, while using $\mathcal{M}=$dark2clear, our approach is able to keep $85.97\%$ accuracy despite the distribution shift of test data. Furthermore, once again, our approach does not sacrifice accuracy on original data (in-distribution) due to the use of OOD detector as an input filter. Figure \ref{fig:benign_OOD} illustrates that $\mathcal{M}=$dark2clear effectively translates dark OOD images (Src-Dist) into its in-distribution equivalent (Target-Dist), that is very similar to the original sample (i.e., deer), which explains its effectiveness to enable more accurate classification on OOD dark data ($+35.64\%$). In conclusion, despite being trained only on clear CIFAR10 data, using our approach, $f$ is able to precisely classify OOD dark test input. Such findings are encouraging to deploy our approach for more critical systems than CIFAR10. For instance, our approach can enable a self-driving car that experienced driving only on roads during the day light to recognize unseen new objects (e.g., roads, traffic signs, etc) during the night.
\begin{table}[t!]

\centering
  \scalebox{.78}{
   \begin{tabular}{|l!{\color{black}\vrule}l!{\color{black}\vrule}l!{\color{black}\vrule}!{\color{black}\vrule}!{\color{black}\vrule}l!{\color{black}\vrule}l!{\color{black}\vrule}}
       \hline
       
&  \multicolumn{2}{c!{\color{black}\vrule}}{\textbf{CIFAR10}}
&  \multicolumn{2}{c!{\color{black}\vrule}}{\textbf{ImageNet}}\\
       \hline
       Test distribution & \textbf{Original} &  \textbf{Darker} & \textbf{Original} &  \textbf{Sharper}\\ 
       \hline
       Baseline model $f$ & 95.19\% &50.33\% &83.58\%&47.18\%\\ 
       \hline
       \shortstack{OOD generalization \\ using $\mathcal{M}=$dark2clear}  & 95.19\% &\textbf{85.97\%}&-&-\\ 
       \hline
       \shortstack{OOD generalization \\ using $\mathcal{M}$=sharp2normal}  &- &-&83.58\%&\textbf{81.34\%}\\ 
       \hline
      
   \end{tabular}}
 
 \vspace*{-.5em}
\caption{Accuracy on benign OOD test data on CIFAR10 (darker images) and ImageNet (sharper images).}
\label{tab:benign_OOD}
\vspace*{-1em}
\end{table}

\textbf{Experiment 2: OOD input due to higher sharpness:}
A test image sharpness may differ from the training images, which could lead to a distribution shift. For instance, higher resolution can lead to a more detailed image with harder edges, making it appear sharper. Table \ref{tab:benign_OOD} shows that the accuracy of the ImageNet model on the same test set, but with higher sharpness, decreases from $83.58\%$ to $47.18\%$. However, using $\mathcal{M}=$ sharp2normal (illustrated in Figure \ref{fig:benign_OOD}), our approach enables an accuracy recovery to reach $81.34\%$ on sharper test data, again reinforcing our observations that our approach not only generalizes on adversarial OOD examples, but also on benign OOD ones as well.

\noindent \fbox{\parbox{.96\linewidth}{\textbf{Conclusion 6:} Our approach not only generalizes on OOD adversarial  examples, but also on OOD benign inputs generated as a result of legitimate distribution shifts.}}

\section{Discussion}\label{sec: discussion}
\vspace*{-1em}
While our approach achieves compelling generalization results for OOD adversarial examples and OOD benign samples, an adaptive adversary may challenge the core insight of our method. Next, we first examine whether such an adversary will succeed, and offer our outlook beyond image-to-image translation and OOD generalization.



\textbf{Adaptive Adversary:}
Consider an adversary that has knowledge of the target model's training distribution and generates in-distribution adversarial examples. If such adversarial examples exist, they may bypass our OOD filter and be directly classified by the model $f$ without passing through our OOD to in-distribution mapping process. We recall that the OOD detector returns an OOD score for each input, and relative to a threshold $\tau$ it determines whether the input requires in-distribution mapping. Consequently, an adaptive adversary may generate samples with an OOD score $<\tau$. One way to achieve this and avoid distribution shift during adversarial perturbations is to make minimal perturbations by allowing a low perturbation size $\epsilon$ (e.g., $\epsilon<0.05$ on CIFAR10).

\begin{figure}[t!]
    
    \centering
    \scalebox{0.5}{
    \includegraphics[width=\textwidth]{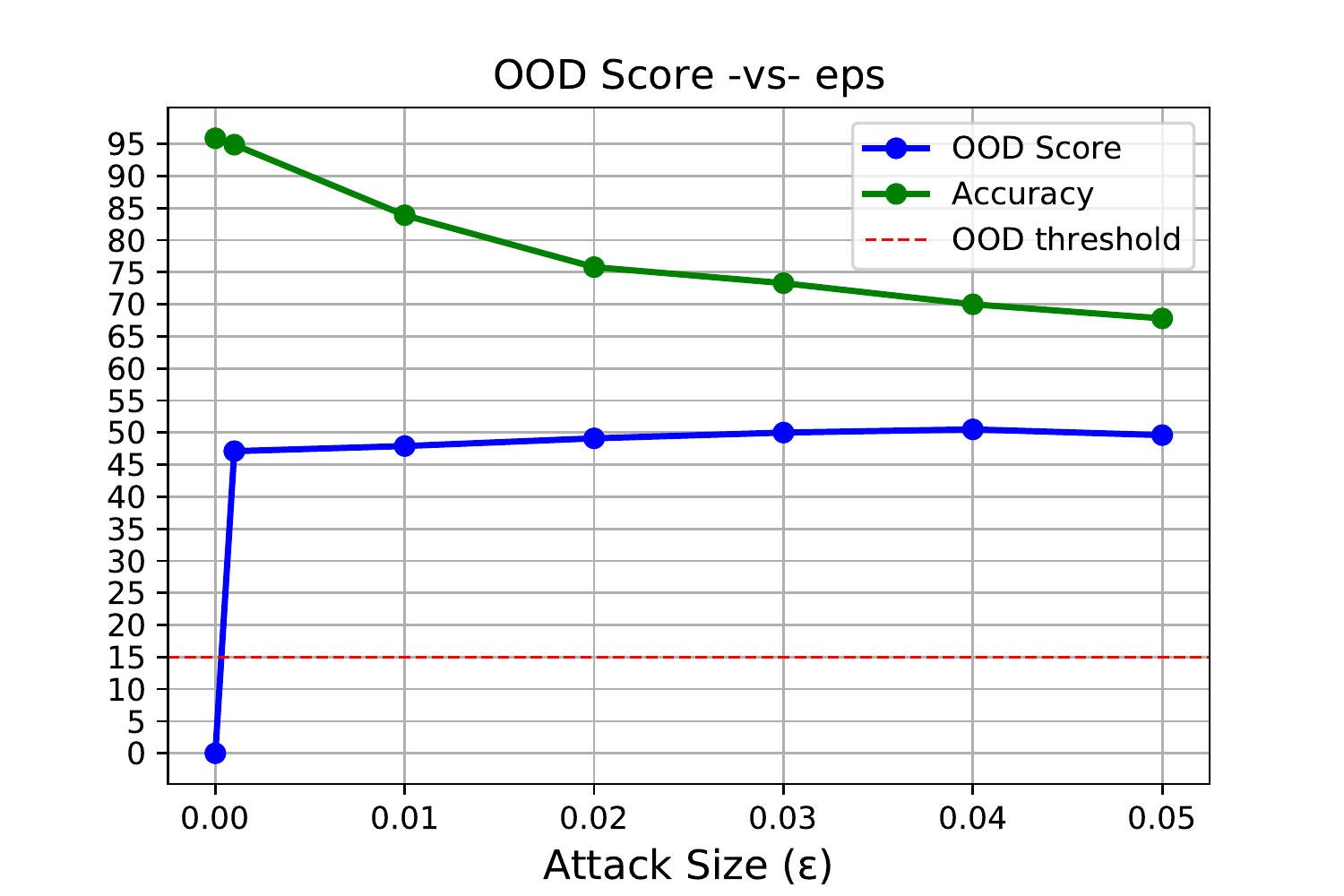}}
   \vspace*{-2em}
    \caption{OOD score and accuracy of CIFAR10 test inputs for lower attack size $\epsilon<0.05$ of the FGSM attack.}
    \label{fig:OOD-vs-eps}
     \vspace*{-1em}
\end{figure}

In order to challenge our defense against such a threat, we play the role of the adaptive adversary and study the OOD scores returned by the OOD detector for lower $\epsilon$ values using FGSM as an input perturbation method. We use 1000 test samples of CIFAR10 to assess the extent to which these samples can break our defense and fool the model. Figure \ref{fig:OOD-vs-eps} shows that even for small perturbations with an attack size as low as $\epsilon=0.001$, the OOD score is still higher than the OOD threshold, which allows our approach to detect such adversarial examples. Additionally, although marginal perturbations with $\epsilon<0.001$ may avoid the distribution shift, Figure \ref{fig:OOD-vs-eps} also shows that it cannot produce evasive examples (i.e., accuracy $> 94\%$). Such observations not only confirm that our approach is resilient against such an attack, but also confirm that distribution shift is a likely to happen after adversarial perturbations which explains their evasive aspect.

\textbf{Beyond Image-to-Image Translation:}
In this paper, we employ Image2Image translation methods to design an OOD generalization method that serves as a defense against adversarial examples in the image classification domain. However, our findings can be leveraged to build more robust defenses in other domains (e.g., malware detection) either by exploring domain-specific formulations of OOD to in-distribution mapping techniques or by turning to the literature of OOD generalization and apply it for the purpose of generalizing on adversarial examples. We observe that solving the OOD problem is a leap forward to solve adversarial examples, due to the cause-effect link between both problems. Hence, we hope future work explores and improves prior OOD generalization methods towards minimizing the IID/OOD generalization gap beyond image classification tasks. 

\textbf{Beyond OOD Generalization:}
Through our extensive experiments on image classifiers, we shed light on the cause-effect link between the OOD problem and adversarial examples problem in order to reach an OOD-generalization-based defense that outperforms the best state-of-the-art defences on three benchmark datasets. We also recognize the fact that the causes of adversarial examples might extend beyond the OOD problem, and OOD generalization alone may not offer a complete solution. ML models must generalize to OOD examples whenever possible. In scenarios where OOD examples do not belong to any known class of a model, other intuitive measures need to be taken. One measure is that when a model fails to bring an OOD input close enough to the natural distribution of a model, it may as well choose to abstain or trigger a ``none of the above'' response. 
\section{Conclusion}\label{sec: concl}
Despite their impressive performance on predictive tasks, ML models struggle when they face OOD inputs: adversarial or otherwise. In this paper we present a framework to systematically study the cause-effect connection between the OOD problem and adversarial examples. After establishing the fact that most adversarial examples happen to be OOD, we leverage image-to-image translation methods to build an OOD generalization framework for image classifiers. 

Through extensive evaluation on three benchmark datasets, we show that our approach consistently outperforms state-of-the-art defenses (adversarial training and ensemble adversarial training) not only in its generalization to OOD adversarial examples but also to OOD benign inputs that result from natural distribution shifts. Contrary to adversarial training based defenses, our approach does not sacrifice accuracy on benign data. We also demonstrate the consistent performance of our approach on alternative designs of the OOD-to-IID mapping approach based on single, mixed, and ensemble of image-to-image translation-based OOD generalization models. Finally, we show how resilient our approach is to an adaptive adversary that strives to constrain adversarial perturbations to produce IID adversarial inputs. 

We hope that this paper paves the way for further research that prioritizes OOD generalization in building trustworthy ML models for high-stakes applications.


\bibliographystyle{acm}
\bibliography{main}

\end{document}